\pdfoutput=1

\documentclass[10pt, journal,twoside]{IEEEtran}

\ifCLASSINFOpdf
\else
\fi

\usepackage{cite}
\usepackage{amsmath}
\usepackage{amsfonts}
\usepackage{algorithm}
\usepackage{algorithmic}
\usepackage{threeparttable}
\usepackage{graphicx}
\usepackage{subfigure}
\usepackage{color}
\usepackage{float}
\usepackage{booktabs}
\usepackage{makecell}
\usepackage{multirow}
\usepackage{pifont}
\usepackage{amssymb}
\usepackage{mathrsfs}

\begin{document}

%
\title{
DCN-T: Dual Context Network with Transformer for Hyperspectral Image Classification
}
%
%
%

\author{Di Wang,~\IEEEmembership{Member,~IEEE,}
        Jing Zhang,~\IEEEmembership{Member,~IEEE,}
        Bo Du,~\IEEEmembership{Senior Member,~IEEE,}\\
        Liangpei Zhang,~\IEEEmembership{Fellow,~IEEE,}
        and Dacheng Tao,~\IEEEmembership{Fellow,~IEEE}
\thanks{\textit{Corresponding authors: Bo Du and Liangpei Zhang.}}
\thanks{D. Wang and B. Du are with the School of Computer Science, Wuhan University, Wuhan 430072, China (e-mail: wd74108520@gmail.com; dubo@whu.edu.cn).}
\thanks{J. Zhang is with the School of Computer Science, 
        Faculty of Engineering, The University of Sydney, Australia (jing.zhang1@sydney.edu.au).}
\thanks{L. Zhang is with the State Key Laboratory of Information
Engineering in Surveying, Mapping and Remote Sensing, Wuhan University,
Wuhan 430079, China (e-mail: zlp62@whu.edu.cn).}
\thanks{D. Tao is with the JD Explore Academy, China and is also with the School of Computer Science, Faculty of Engineering, The University of Sydney, Australia (dacheng.tao@gmail.com).}
}

%
%

\markboth{Journal of \LaTeX\ Class Files,~Vol.~14, No.~8, August~2015}{Wang \MakeLowercase{\textit{et al.}}: DUAL CONTEXT NETWORK WITH TRANSFORMER FOR HSI CLASSIFICATION}
%



\maketitle

\begin{abstract}
  Hyperspectral image (HSI) classification is challenging due to spatial variability caused by complex imaging conditions. Prior methods suffer from limited representation ability, as they train specially designed networks from scratch on limited annotated data. We propose a tri-spectral image generation pipeline that transforms HSI into high-quality tri-spectral images, enabling the use of off-the-shelf ImageNet pretrained backbone networks for feature extraction. Motivated by the observation that there are many homogeneous areas with distinguished semantic and geometric properties in HSIs, which can be used to extract useful contexts, we propose an end-to-end segmentation network named DCN-T. It adopts transformers to effectively encode regional adaptation and global aggregation spatial contexts within and between the homogeneous areas discovered by similarity-based clustering. To fully exploit the rich spectrums of the HSI, we adopt an ensemble approach where all segmentation results of the tri-spectral images are integrated into the final prediction through a voting scheme. Extensive experiments on three public benchmarks show that our proposed method outperforms state-of-the-art methods for HSI classification. The code will be released at https://github.com/DotWang/DCN-T.

\end{abstract}


%
\IEEEpeerreviewmaketitle

\section{Introduction}
%
%
%
%

\IEEEPARstart{H}{yperspectral} images (HSIs) offer a unique advantage over other forms of data, as they can accurately describe the physical characteristics of objects. By utilizing advanced sensors, HSIs possess abundant spectral information characterized by hundreds of bands, obtained by receiving electromagnetic signals in dense and narrow wavelength ranges. Consequently, HSIs have found extensive application in various recognition tasks, such as precision agriculture \cite{agriculture_1,zhang2020empowering} and environmental monitoring \cite{env_monit_1}. These tasks are typically realized by implementing a pixel-wise classification approach.

HSI suffers from a significant challenge referred to as ``spatial variability'', wherein the same object exhibits varying characteristics in different regions. This is attributed to factors such as atmospheric interference \cite{atmosphere_1}, light angle \cite{angle_1}, and other such effects that occur during the imaging process. The resulting problem of mixed categories can lead to inter-class similarity, i.e., similar spectral signatures are presented by different categories, thereby increasing the difficulty of classification.

To address the above issue, it is necessary to extract discriminative features. As convolutional neural networks (CNNs) have emerged as the most widely-used network in deep learning-based models for HSI classification, researchers have designed various network structures, including overall framework construction \cite{bassnet,Lee2017TIP,dffn,PyramidCNN,hang_cmcl_tgrs_2022,caa_grsl,sslr_tip} and effective module utilization \cite{rlnet,ssatt,ssan,specattennet}. Some approaches have also used automatic network architecture search algorithms \cite{autocnn_hsi,3danas,sstn}. Additionally, recurrent neural networks \cite{DirectRNNinHSI,SS_CLSTM,SS_LSTM,CascadedRNNinHSI} and graph convolutional neural networks \cite{mdgcn,cadgcn,cegcn,minigcn,wfcg_tip} have also been explored. Typically, these networks are trained from scratch on HSI.

Numerous studies \cite{asr_review,zhang2022vitaev2,wang_rsp_2022,wang_vitrvsa_2022} have demonstrated the effectiveness of using pretrained weights for enhancing the performance and convergence of deep networks. There are many off-the-shelf networks pretrained on a large-scale dataset, e.g., ImageNet~\cite{deng2009imagenet} containing millions of images from the real world. They have obtained an excellent ability in extracting discriminative features. However, these networks are not directly applicable to HSI classification due to the disparity of channel numbers between natural images and HSIs. A naive solution is transforming all spectral channels into a tri-spectral image or selecting only three channels, which however will lead to a significant information loss compared to the original abundant spectrums. Besides, generating a large number of tri-spectral images through random sampling and evaluating in an ensemble fashion is computationally infeasible. Specifically, for an HSI with $L$ channels, the maximum number of tri-spectral images that can be produced is $P_3^L=L(L-1)(L-2)$, where $L$ is usually greater than 100.

We believe that appropriate tri-spectral images can be generated by selecting specific channels for dimensionality reduction while retaining as much valuable information as possible, and this approach enables the use of pretrained networks. To this end, we propose a novel tri-spectral image generation pipeline that converts the HSI into a series of high-quality tri-spectral images. Specifically, we divide all channels into several groups, and each group is responsible for one of the channels in the generated tri-spectral images. Each tri-spectral image can be produced by combining three randomly selected arbitrary groups. The grouping operation aims to balance accuracy and efficiency, allowing us to explore a suitable ratio for dimension reduction. The random selection guarantees that the preferences of specific categories can be satisfied and improves the diversity of the generated images. Additionally, the combination of different groups considers the wavelength order to maintain original information. Finally, this specially designed pipeline generates numerous tri-spectral images to form a tri-spectral dataset, akin to RGB images used during pretraining (detailed in section III-A).

The existing literature \cite{wang_rsp_2022} suggests that the universal representations derived from ImageNet pretraining can be beneficial in remote sensing tasks. Therefore, once the original HSI is converted to tri-spectral images, we can employ off-the-shelf ImageNet pretrained backbones for HSI classification. Our focus is primarily on extracting spatial contexts, which have been demonstrated to be useful for HSI classification in previous studies \cite{rs_amcnn,msage_cal,zhang2018tipforHSIC}. Conventional spatial feature-based HSI classification methods primarily rely on exploiting neighboring information in patches surrounding the target pixel. However, the obtained information is usually restricted in a limited range, which is closely related to the window size. As a result, image-level methods have emerged, in which the entire image is directly input into the network to predict the categories of all pixel locations simultaneously \cite{freenet,fcontnet,fcsn_tip,hang_mpsn_tgrs_2022}, similar to the segmentation task in computer vision (CV). Nonetheless, convolutions are local operations that are inefficient at capturing long-range contexts. Then, the non-local self-attention (SA) mechanism is utilized in \cite{sacn_adversarial_tip,lw_sa_rn,rssan,h2an} to address this problem. However, it can bring about high computational overheads. Hence, finding an efficient way to capture useful context for improving HSI classification is a critical problem that needs to be addressed.

\begin{figure}[t]
  \centering
  \includegraphics[width=0.6\linewidth]{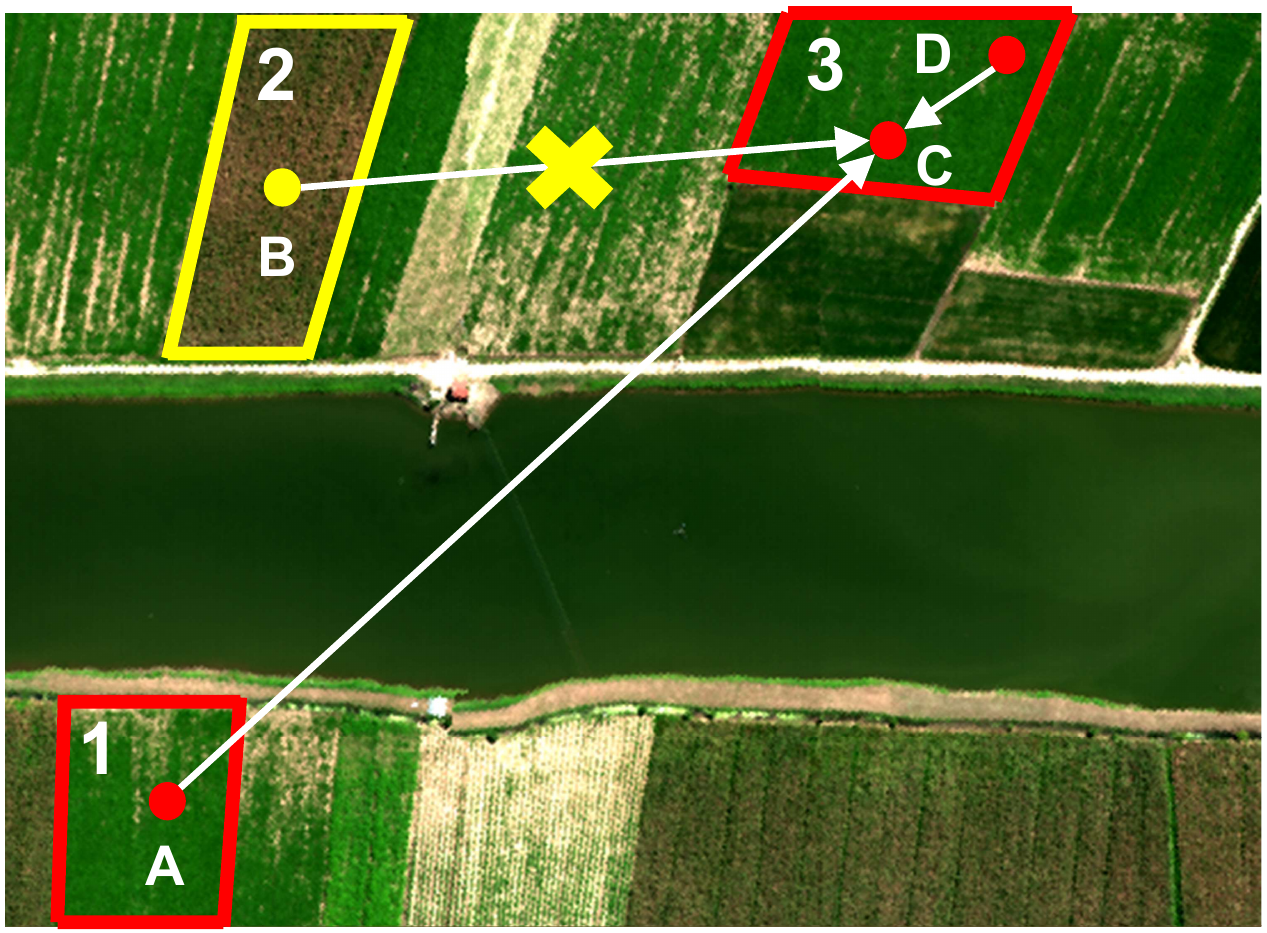}\\
  \caption{Illustration of three homogeneous areas. Areas $1$ and $3$ marked by the red box have similar characteristics. Within area 1 and 2, points $A$ and $B$ are separated, whereas both points $C$ and $D$ are located in area 3. This image is generated using the $7$-th, $5$-th, and $3$-rd groups on the LongKou scene of the WHU-Hi dataset to create three channels, which are analogous to the red, green, and blue channels. The process of tri-spectral image generation will be discussed in detail in section III-A.
  }
  \label{area}
\end{figure}

The context of a reference position can be represented by the pixels in a position set. For natural image segmentation, some methods \cite{ocrnet,acfnet,cdgcnet} use the pixels in the same category to recognize the position set, which however is infeasible for HSI since the available training samples lack sufficient annotations (see Figure \ref{dataset}). Nevertheless, in HSI, there exist many homogeneous areas around the reference pixel, with similar pixel values and specialized semantic and geometric properties (see red boxes in Figure \ref{area}). The semantic property means internal pixels tend to be in the same category, while the geometry property indicates that the boundaries of these areas match object contours. As a result, the homogeneous area can provide valuable spatial context information that can be leveraged to enhance classification performance. Since the homogeneous areas of the same category may not be connected, it is necessary to aggregate the context information across different areas after capturing the intra-area context. Figure \ref{area} presents an example, where the reference point $C$ should perceive the information from location $D$ and $A$, while pixel $B$ should not be involved. To fully exploit the features within these homogeneous areas, we leverage an advanced transformer \cite{selfattention} that contains a multi-head SA (MHSA) mechanism. It can calculate similarities between pixels in multiple subspaces, thereby capturing more diverse contexts. Technically, after transforming the HSI into tri-spectral images following the proposed pipeline and using the ImageNet pretrained backbone to obtain corresponding features, we conduct similarity-based clustering on neighboring pixel representations to obtain homogeneous areas. Then, we develop a transformer-based dual context module (DCM) to extract two types of context, i.e., regional adaptation context (RAC) and global aggregation context (GAC), which correspond to intra-area and inter-area context, respectively. We call the whole network to dual context network with transformer (DCN-T).

As previously discussed, each tri-spectral image is produced by selecting specific channels from the original HSI, thereby containing only a subset of the spectral information. As the original HSI consists of numerous channels, the segmentation results obtained from different tri-spectral images can be combined. To achieve this, we propose a voting scheme that enables the fusion of the segmentation maps to improve classification accuracy. To sum up, the main contribution of this paper is three-fold.

\begin{itemize}
  \item [1)] We design a novel pipeline for generating high-quality tri-spectral images from HSI, which allows the use of off-the-shelf ImageNet pretrained backbone for extracting discriminative features and addressing the issue of spatial variability.
    \item [2)] We develop an end-to-end segmentation network named DCN-T. Specifically, we introduce a clustering scheme for contour-adaptive homogeneous area generation and adopt the transformer to capture both regional adaptation and global aggregation spatial contexts that are separately lying in or between these areas.
    \item [3)] We propose a new voting scheme that effectively harnesses the rich spectrums of the HSI by integrating the segmentation results of all tri-spectral images, thereby improving the classification accuracy. Extensive qualitative and quantitative experiments on three public HSI benchmarks demonstrate that the proposed method outperforms state-of-the-art methods.
\end{itemize}

The remainder of this paper is organized as follows. Section II introduces the related works. Section III presents the tri-spectral image generation pipeline and the DCN-T model. Experiment results and analyses are shown in section IV. Finally, Section V concludes the paper.

\section{Related Work}

\subsection{Dimensionality Reduction in HSI Classification}
Besides spatial variability, the HSI has another classical issue that is named the curse of dimensionality or the Hughes phenomenon \cite{Xia2016}, due to the abundant bands. This issue is usually caused by the unbalance between limited samples and high-dimensional features, easily leading to overfitting. This issue implicitly implies that the channels in HSI tend to be redundant. The most intuitive idea to address this issue is to conduct a reasonable dimensionality reduction.

Traditionally, there are two paradigms to reduce dimension. The first one is feature selection, which aims to select the most discriminative bands by some manually designed criteria \cite{feature_select_1}. However, the spectral redundancies of different objects usually take place in different channels, increasing the difficulty of band selection. In addition, in these methods, except for the chosen bands, other channels are unavoidably discarded, though they may be still valuable. Another one is feature extraction, where the original high-dimensional features are directly projected to a low-dimensional space, such as the principal component analysis \cite{PCAinHSIclassify}. Nevertheless, spectral information is inevitably lost.

Considering the above challenges, most deep learning-based pixel-level classification methods usually use all bands when extracting spectral feature \cite{hu2015deep}, e.g., a patch of size $s \times s \times L $ centering at the reference pixel is fed into the network \cite{3dcnn,ssrn}. Here, $s$ is the height and width of the patch, and $L$ is the number of HSI bands. There are also some methods to reduce the channel dimension \cite{ssan_rs,assmn,ssun}. However, they still conduct the patch-level classification by providing the network with the patches in size of $s \times s  \times p$, where $p$ is the number of channels after dimensionality reduction. In recent image-level classification or segmentation-based methods, the whole image is fed into the network \cite{ssfcn,enl_fcn,dasgcn}, where the input data contain all channels, leading to a huge computational cost. Compared with the aforementioned methods, we propose to transform the original HSI into a series of tri-spectral images, achieving a better trade-off between dimensionality reduction and maintaining the integrity of spectral information.

\subsection{CNN-based HSI Classification Methods}

Due to the excellent ability in local perception and computational efficiency of convolutions, many CNN-based methods have emerged in the HSI classification field.  Initially, pixel vectors or patches are fed into the network.\cite{hu2015deep,3dcnn,ssrn,ssun,assmn}. For example, \cite{hu2015deep} constructs a five-layer 1-D CNN to extract spectral features. 2-D and 3-D convolutional kernels are separately adopted in \cite{3dcnn} to obtain spatial or spectral-spatial features. \cite{ssrn} first uses CNN to capture spectral features along the channel direction and then aggregates the spatial information. While \cite{Li2017} uses CNN to extract discriminative representations from designed pixel pairs. Besides the above pixel-level or patch-level classification approaches, recently, many image-level (a.k.a., patch-free) methods have been proposed recently \cite{ssfcn, enl_fcn, fcontnet, fcsn_tip}. Among these networks, the fully convolutional networks (FCN) \cite{FCN} --- a special family of CNN, has been widely used, where the output is the same size as the input. Based on the FCN, \cite{ssfcn} introduces dilation convolution to enlarge the receptive fields, while \cite{freenet} adopts an encoder-decoder structure to progressively recover details. However, since convolutions are local operations, CNN is not good at extracting long-range contextual information to obtain global perception, which is important for HSI classification.

\subsection{Attention-based HSI Classification Methods}

The attention mechanism, which aims to mimic human vision that imposes different importance on different areas of a scene, has been frequently used in HSI classification to enhance features. This mechanism can be flexibly combined with other components. \cite{rssan, ssatt} directly adopt classical attention modules such as squeeze-and-excitation \cite{se_block} or convolutional block attention module \cite{cbam} to weight different channels or spatial positions. \cite{ssan_rs} defines a group of trainable parameters to merge the features from different branches through weighted addition. Different from the above explicit feature enhancement ways, the SA mechanism is used to implicitly enhance features since it can effectively capture the relationships between different locations. For example, \cite{ssan} directly employs the non-local SA module \cite{non-local} to obtain spatial contexts, while a more light SA module named criss-cross attention \cite{ccnet} is utilized in the image-level network \cite{enl_fcn}. In addition, the graph attention is employed to process irregular spatial structure of the HSI \cite{wfcg}. On the foundation of the SA mechanism, a more powerful variant, MHSA, has been proposed. The MHSA simultaneously conducts different SA in multiple subspaces such that more abundant contexts can be captured, leading to more effective feature representations. The MHSA is the core component of the transformer, which will be introduced next.

\subsection{Transformer-based HSI Classification}

The transformer is first proposed in \cite{selfattention} for machine translation and then achieves great success in the NLP field \cite{bert}. Although the SA mechanism has been applied to the CV field \cite{non-local,danet,ccnet}, the transformer will not be successfully applied for image understanding until the end of 2020 \cite{vit,swint,xu2021vitae,pvt,pvt_v2,zhang2022vitaev2,zhang2022vsa,carion2020end,wang2021exploring,wang2021fp}. Recently, the transformer has drawn increasing attention in the HSI classification community. The first work is \cite{hsi_bert}, where the BERT \cite{bert} is directly adopted on the flattened input patches by analogizing each pixel as a ``word vector'' in NLP. Then \cite{spectralformer} and \cite{tsst} adopt the transformer to extract spectral context, where the feature embedding is generated by taking one or multiple channels of the input patches. The difference is that \cite{tsst} uses the transformer on the extracted features from CNN, while \cite{spectralformer} adopts a cascade scheme where multiple transformer encoders are stacked. \cite{ssftt} and \cite{lsfat} use transformers to obtain spatial contexts through the flattened feature maps that are extracted by 2-D or 3-D CNN. Different from these patch-level classification approaches, we propose a novel DCN-T, which is an image-level segmentation network. A very recent work \cite{lessformer} adopts multiple transformer encoders to separately extract spatial and spectral contexts after directly splitting the features of the whole image. Different from it, we not only use transformer encoders to capture the contexts within and between homogeneous areas for obtaining a discriminative spatial feature, but also employ a transformer decoder to further refine the feature by aggregating the information of these areas. \cite{tsst} separately converts each single-band input to three channels through a trainable linear projection layer to utilize ImageNet pretrained networks. However, the generated three channels are highly correlated, i.e., the result can be regarded as a simple stacking of three single-band images, affecting the performance of extracted features. In this paper, we transform the HSI into a series of high-quality tri-spectral images with a well-designed pipeline to match the input size of ImageNet pretrained networks. Then, based on the features extracted by the backbone network from these tri-spectral images, we use the transformer to further extract the regional and global context in and between homogeneous areas, greatly improving the representation ability of features and the HSI classification performance.

\section{Method}

In this section, we first introduce the designed tri-spectral image generation pipeline. Then, we describe the details of the proposed transformer-based DCM. Combining the DCM with the ImageNet pretrained backbone network, we obtain the DCN-T for HSI classification.

\subsection{Tri-spectral Image Generation}

The tri-spectral image generation pipeline includes three steps: grouping and aggregation, random band selection, and image stretching.

\subsubsection{Grouping and Aggregation}

We first transform each HSI into several sub-images by splitting its channels equally. For example, for an HSI $\mathbf{X}_{HSI}=\{x_1,\cdots,x_L\} \in \mathbb{R}^{H \times W \times L}$, where $H,W$ and $L$ are height, width and the number of channels, and each group of a sub-image has $L/G$ channels, where $G$ is the number of groups. Since the adjacent channels are highly correlated \cite{DirectRNNinHSI,ssun,assmn,ssan_rs}, for the convenience of the subsequent aggregation, we sequentially split the channels along the spectral direction instead of random grouping. Now the HSI can be symbolized as $\mathbf{X}_{HSI}=\{X_1,\cdots,X_{G}\}$, and the $i$-th group is $X_i=\{x_{(i-1) \frac{L}{G}+1},\cdots,x_{i \cdot \frac{L}{G}}\}$.

Then, we conduct the dimensionality reduction by aggregating the channels in each group to tackle the Hughes phenomenon issue. Concretely, we calculate the mean value of these channels, and each sub-image now becomes a 2-D map, which will be used as a channel to construct a tri-spectral image. Concretely, for the $i$-th group, the 2-D map is calculated as:
\begin{equation}
  \widetilde{X}_i = \frac{G}{L} \sum\limits_{j=1}^{L/G} x_{(i-1) \frac{L}{G}+j},
\end{equation}
where $\widetilde{X}_i \in \mathbb{R}^{H \times W}$, and the original HSI is transformed to $\widetilde{\mathbf{X}}_{HSI}=\{\widetilde{X}_1,\cdots,\widetilde{X}_{G}\}$.

\subsubsection{Random Band Selection}

  \begin{figure}[t]
    \centering
    \includegraphics[width=1\linewidth]{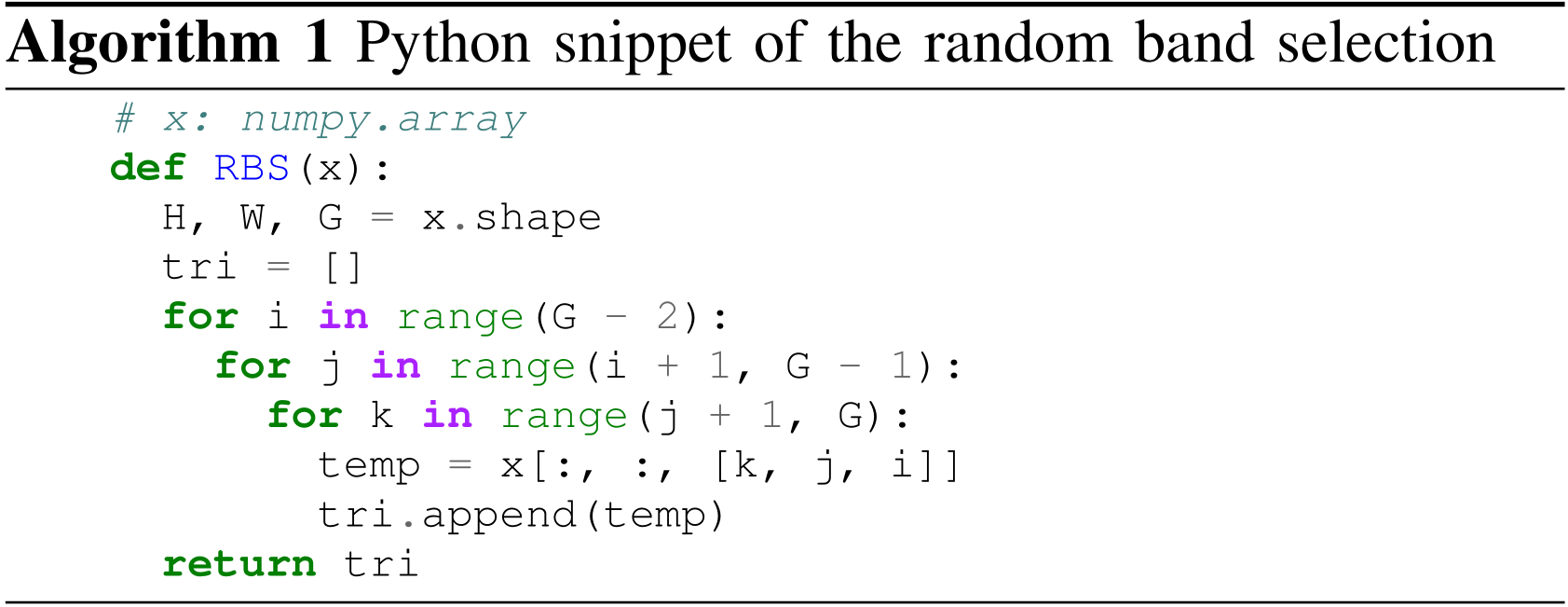}\\
  \end{figure}

Any three sub-images in $\widetilde{\mathbf{X}}_{HSI}$ can be used to generate a tri-spectral image. Since different categories may prefer different bands, i.e., be better reflected by different bands, we randomly select three sub-images as the corresponding channels. In addition, inspired by the channel order in RGB images, we further consider the wavelength of each band to determine the relative order of the channels in the generated images. Specifically, we adopt the sub-image that corresponds to the maximum wavelength as the first channel, the medium one as the second channel, and the minimum one as the last channel. This simple strategy not only maintains the original spectral information, but also further excludes unnecessary combinations and greatly reduces the number of generated tri-spectral images. In this way, the set of the obtained tri-spectral images can be denoted as:
\begin{equation}
  \mathbf{X}_{TRI}^{0} = RBS(\widetilde{\mathbf{X}}_{HSI}).
\end{equation}
Here, $RBS(\cdot)$ denotes the random band selection procedure, whose details are shown in Algorithm 1. The randomness is reflected by the fact that three channels in the tri-spectral image are not necessary to be adjacent in $\widetilde{\mathbf{X}}_{HSI}$. $\mathbf{X}_{TRI}^{0} = \{I_1^{0},\cdots,I_M^{0}\}$, where $I_i^{0}, i=1,\cdots,M$ is the obtained initial tri-spectral image, which will be enhanced later. $M$ is the number of the generated tri-spectral images, i.e.,
\begin{equation}
  M = \binom{G}{3}=\frac{1}{6}G(G-1)(G-2).
\end{equation}
Also, we can say that capacity of the generated tri-spectral image set is $M$.

\subsubsection{Image Stretching}

The obtained initial tri-spectral image $I^{0}$ is blurry. We adopt the linear 2\% stretching, which is a popular and effective stretching method that has been widely used, such as in the ENVI software, for preprocessing remote sensing images to generate high-contrast images. The reason why we use this stretching is although $I^{0}$ is a tri-spectral image like the natural image, it still depicts remote sensing scenes.

The linear 2\% stretching scales the values between the 2\% and 98\% positions in the cumulative histogram of all channels to [0, 255], while values less than 2\% or greater than 98\% sites will be directly assigned to 0 and 255. This procedure is can be described as:
\begin{equation}
  I=LTS(I^{0}), \quad LTS(x)=
  \begin{cases}
    \cfrac{x-p}{q-p} \cdot 255 &  p \leq x \leq q \\
    0 & x > p \\
    255 & x < q    \\
  \end{cases},
\end{equation}
where $p,q$ are 2\% and 98\% positions in the cumulative histogram of image $I^{0}$. $LTS$ represents the linear 2\% stretching, and $I$ is the finally generated high-quality tri-spectral image.

\begin{figure}[t]
  \centering
  \includegraphics[width=0.7\linewidth]{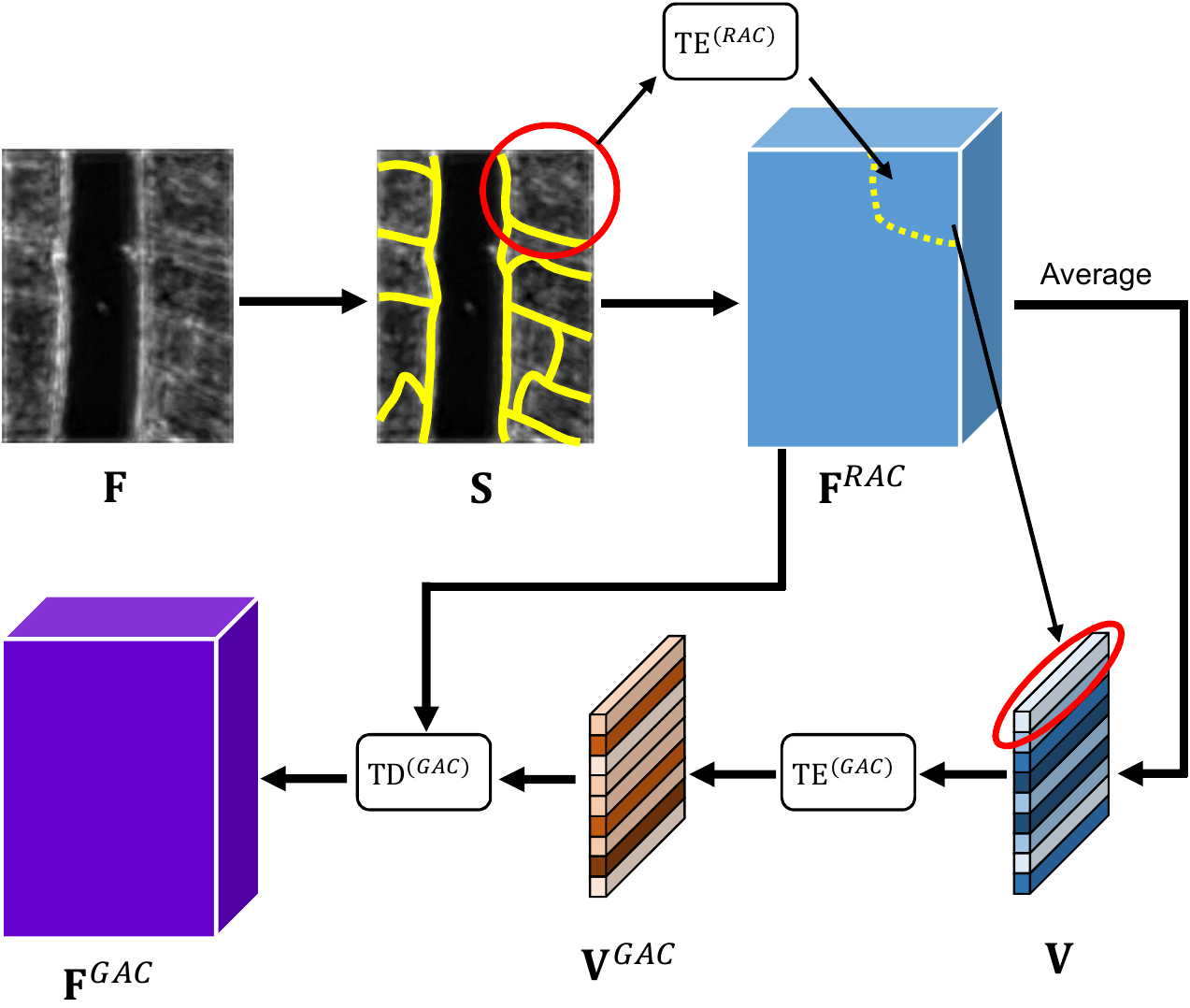}\\
  \caption{Implementation details of DCM. The homogeneous areas $\mathbf{S}$ are adaptively generated on the input feature $\mathbf{F}$. Then, each area in $\mathbf{S}$ goes through a transformer encoder $\text{TE}^{(RAC)}$ to capture the RACs. A group of descriptors $\mathbf{V}$ are produced by separately averaging the pixel representations of each area on the obtained $\mathbf{F}^{RAC}$. After another transformer encoder $\text{TE}^{(GAC)}$, $\mathbf{V}$ is transformed to $\mathbf{V}^{(GAC)}$, where the GACs are extracted. At last, using $\mathbf{V}^{(GAC)}$ and $\mathbf{F}^{RAC}$, a transformer decoder $\text{TD}^{(GAC)}$ is adopted for feature reconstruction. The red circle represents a homogeneous area and its corresponding descriptor that is obtained by a transformer encoder with an averaging operation.}
  \label{dcm}
\end{figure}

\subsection{Dual Context Module}

In this section, we first introduce the structure of the transformer for context capturing in the proposed DCM. Then, we present details of the DCM, which is shown in Figure \ref{dcm}, including homogeneous areas generation and the extraction of RAC and GAC.

\subsubsection{Transformer}

\begin{figure}[t]
  \centering
  \includegraphics[width=0.7\linewidth]{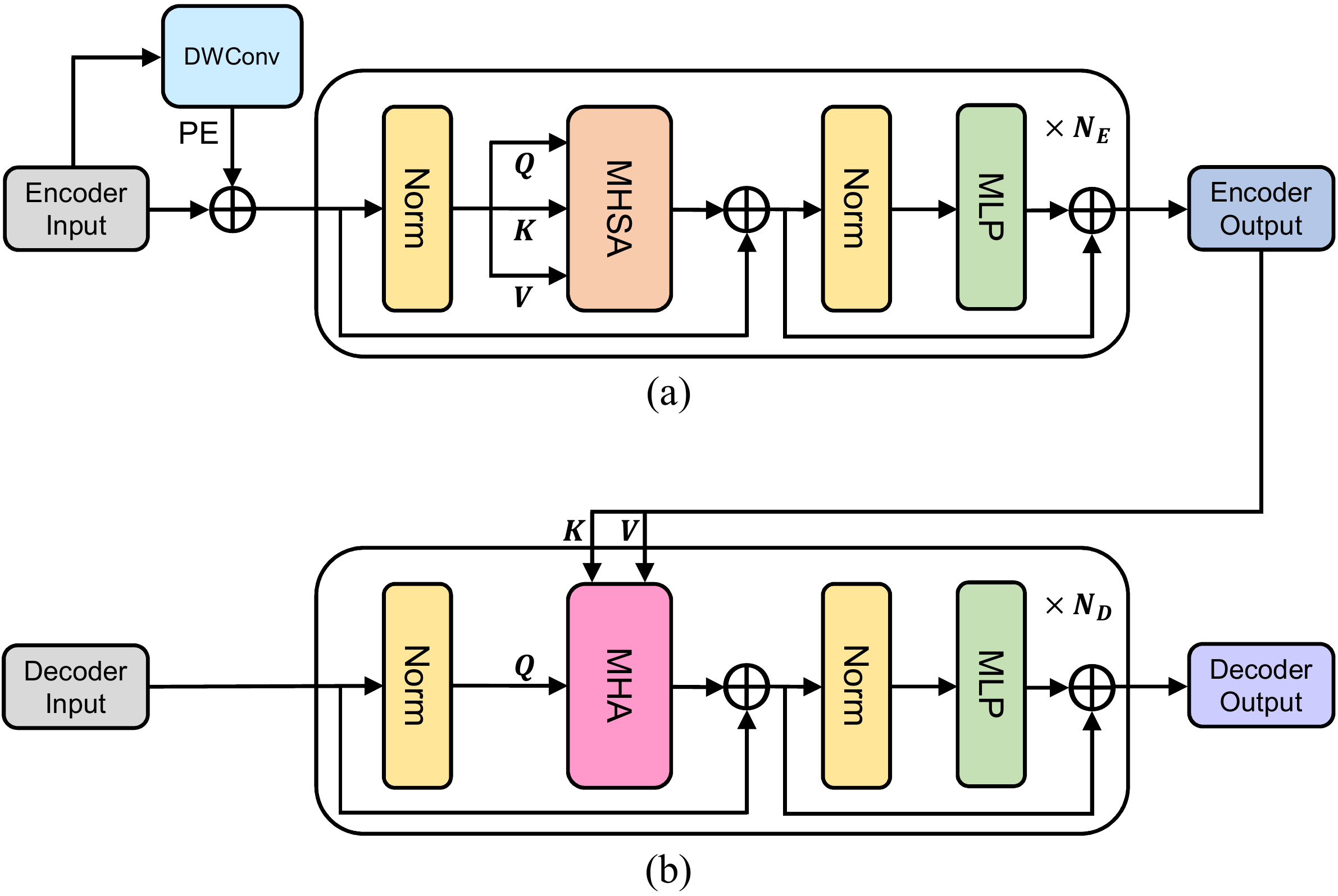}\\
  \caption{Structure of the adopted transformer. (a) Encoder. (b) Decoder.}
  \label{transformer}
\end{figure}

Figure \ref{transformer} shows the structure of the used transformer. It contains an encoder and a decoder, whose number of layers is controlled by $N_E$ and $N_D$, respectively. Notice that compared with the original transformer in \cite{selfattention}, we remove MHSA in the decoder part. Since the input of this decoder is the generated RAC features, where the MHSA operations have been implemented. Thus, it is unnecessary to conduct them again. The details will be introduced later.

\noindent\textbf{Encoder.} Assume the input of the encoder is $X_{EI}$, considering the unequal length tokens that are obtained by the generated irregular homogeneous areas, we will not adopt the predefined positional encoding for fitting with the size of the embedding features. Following \cite{pvt_v2}, we use a 3 $\times$ 3 depthwise convolution (DWConv), which is a group convolution whose group number equals channels to produce the positional encoding. Thus, according to Figure \ref{transformer} (a), the procedure of the transformer encoder can be formulated as:
\begin{equation}
  \begin{split}
    \text{TE}(X_{EI},P) = \text{Res}(&\text{MLP}(\text{Norm}(\cdot)),\\
  &\text{Res}(\text{MHSA}(\text{Norm}(\cdot)), X_{EI}+P)),
  \end{split}
\end{equation}
where $\text{TE}$ is an abbreviation of transformer encoder. $P$ is the positional encoding, which can be computed using $X_{EI}$. $\text{Res}(\cdot)$ represents a residual connection function. MLP and Norm represent the multi-layer perceptron and the layer normalization, respectively.

MHSA is the key part of the transformer which performs SA multiple times in parallel to obtain diverse attention maps in different projected subspaces. The SA mechanism can capture the internal relationships in the input feature $X$ by receiving a triplet $Q, K, V$, where
\begin{equation}
  Q = W_Q^T X, \quad  K = W_K^T X, \quad  V = W_V^T X .
\end{equation}
Here, $Q, K, V$ are separately called the query, key, and value, $W_Q, W_K, W_V \in \mathbb{R}^{C \times d}$ are three learnable weight matrices, $C $ is the channel number of $X$. The $\text{SA}$ is implemented as:
\begin{equation}
  \text{SA}(Q,K,V) = \sigma(\frac{Q^TK}{\sqrt{d}})V^T,
\end{equation}
where $d$ is the channel number of the element in the triplet, and $\sigma$ represents the softmax function. The $\text{SA}(Q, K, V)$ is called the output of a $head$. MHSA firstly concatenates these outputs in channel dimension and then maps them to the original channel number $C$, i.e.,
\begin{equation}
  \text{MHSA}(X) = W_O \text{Concat}(head_1,\cdots,head_h)^T.
\end{equation}
Here, $head_i = \text{SA}(Q_i,K_i,V_i)$, where $Q_i = (W_Q^i)^T X, K_i = (W_K^i)^T X,V_i = (W_V^i)^T X$. $W_Q^i, W_K^i, W_V^i \in \mathbb{R}^{C \times d}$ and $W_O \in \mathbb{R}^{C \times hd}$ is a linear transformation matrix, $h$ is the number of heads. In our implementation, $hd=C$ for convenience.

\noindent\textbf{Decoder.} As shown in Figure \ref{transformer} (b), the procedure of the transformer decoder is formulated as:
\begin{equation}
  \begin{split}
    \text{TD}(X_{DI}, X_{EO}) = \text{Res}(&\text{MLP}(\text{Norm}(\cdot)),\\
  &\text{Res}(\text{MHA}(\text{Norm}(\cdot), X_{EO}), X_{DI})),
  \end{split}
\end{equation}
where $\text{TD}$ is the abbreviation of transformer decoder, $X_{DI}$ is the another required input, while $X_{EO} = \text{TE}(X_{EI})$ is the output of the encoder. Compared to the encoder, the decoder has no positional encoding, and the MHSA is replaced by the multi-head attention (MHA). Compared with the MHSA which only needs a single input, the MHA has different inputs. Concretely, the query is from feature $X_1$, while the key and value are from another feature $X_2$. Thus, the MHA can be formulated as:
\begin{equation}
  \text{MHA}(X_1, X_2) = W_O \text{Concat}(head_1,\cdots,head_h)^T,
\end{equation}
where $head_i = \text{SA}(Q_i,K_i,V_i), Q_i = (W_Q^i)^T X_1, K_i = (W_K^i)^T X_2,V_i = (W_V^i)^T X_2, W_Q^i, W_K^i, W_V^i \in \mathbb{R}^{C \times d}$ and $W_O \in \mathbb{R}^{C \times hd}$. Through the decoder, the feature dimension is reprojected to $C$.

\subsubsection{Homogeneous Area Generation}

Assume the input feature is $\mathbf{F}=\{f_1,\cdots,f_N\}\in\mathbb{R}^{C\times H_1 \times W_1}$, here $N= H_1 W_1$, $C$ is also the number of the input channels of the subsequent transformer. One straightforward idea to obtain homogeneous areas is to group similar pixel representations, which are usually spatially adjacent. Concretely, $\mathbf{F}$ is first split into $Z$ regular grids $\mathbf{S}^{(0)}=\{S_1^{(0)},\cdots,S_Z^{(0)}\}$, where the size of each grid is $s \times s$. For the $i$-th grid $S_i^{(0)}$, it should locate in the range of $[(n_h-1)\cdot s:n_h\cdot s,(n_w-1)\cdot s:n_w \cdot  s]$, where $n_h=\lfloor (i-1)/(W_1 / s) \rfloor +1, n_w  = \text{mod}(i-1,W_1 / s)+1$ and $s=\sqrt{N/Z}$, $i=1,\cdots,Z$.

Then, we average these grids to initialize a series of cluster centers, which could be formulated as:
\begin{equation}
  r_i^{(0)} = \frac{1}{s^2} \sum\limits_{j\in S_i^{(0)}} f_j.
\end{equation}
where $r_i^{(0)}$ is the $i$-th initial cluster center. $j$ is a position in the $i$-th initial homogeneous area $S_i^{(0)}$, corresponding to the pixel representation $f_j$.

The homogeneous areas are obtained iteratively. Specifically, in $t$-th $(t=1,\cdots,T)$ iteration, we measure the similarity between each pixel and the cluster centers. For pixel $j$ and cluster $i$ we have:
\begin{equation}
  A_{ji}^{(t)} = \exp(-D(f_j,r_i^{(t-1)})),
  \label{comp_a}
\end{equation}
where $A\in\mathbb{R}^{N \times Z}$ is an affinity matrix, $A_{ji}^{(t)}$ is the value of the $j$-th row and $i$-th column in $A$ at the iteration $t$, $D$ is a distance metric. In this paper, the square euclidean distance is adopted. Thus, $D(x_1,x_2) = ||x_1-x_2||^2$. Based on $A$, each pixel is assigned to a new cluster:
\begin{equation}
  J_j^{(t)} = \mathop{\arg\max}_i A_{ji}^{(t)},
\end{equation}
$J_j^{(t)}$ is the index of the cluster to which pixel $j$ is assigned in the iteration $t$ and the new cluster centers are obtained as:
\begin{equation}
  r_i^{(t)} = \frac{1}{n_i^{(t)}} \sum\limits_{j\in S_i^{(t)}} f_j.
\end{equation}
Here, $S_i^{(t)}$ is an area where its inner position $j$ satisfies $J_j^{(t)} = i$, and can be represented as $S_i^{(t)} = \mathcal{P}(J_k^{(t)} = i,k=1,\cdots,N)$, meaning $S_i^{(t)}$ is a position set that is constructed when the condition in the bracket is satisfied, where $\mathcal{P}$ is also a position set that contains all spatial positions. $n_i^{(t)}$ is the number of pixels included in $S_i^{(t)}$.

However, since the argmax operation is non-differentiable. The above procedure can not be incorporated into the network for end-to-end training. To address this issue, we implement a soft assignment \cite{ssn_eccv}, and the cluster centers are obtained by
\begin{equation}
  r_i^{(t)} = \frac{\sum_{j=1}^N A_{ji}^{{(t)}} f_j}{\sum_{j=1}^N A_{ji}^{{(t)}}}.
\end{equation}

Finally, after the $T$-th iteration, the homogeneous areas $\mathbf{S}^{(T)} = \{S_1^{(T)},\cdots,S_Z^{(T)}\}$ can be produced through

  \begin{equation}
  \begin{split}
    S_i^{(T)} =  \mathcal{P}( & J_k^{(T)}  = i, k=1,\cdots,N), \\  
    J_j^{(T)}  &= \mathop{\arg\max}_i A_{ji}^{(T)}.
  \end{split}
  \end{equation}

In practice, not all clusters are involved in the computation of $A$. In the formula (\ref{comp_a}), for each pixel, only the surrounding clusters are considered to reduce unnecessary calculations.

\subsubsection{Regional Adaptive Context} RACs are captured on the generated homogeneous areas $\mathbf{S}$ (The superscript $T$ is omitted for convenience). The ``Adaptive'' means these homogeneous areas are adaptively obtained, which are dynamically adjusted during network training to continuously fit the contours of objects in $\mathbf{F}$.

Since RACs are the relationships between internal pixels in each homogeneous area $S_i$, these relationships can be directly modeled by the attention mechanism. We only use the transformer encoder to encode the RACs. Concretely, one encoder is used to separately process each homogeneous area. For the $i$-th area, the obtained RAC is:
\begin{equation}
  \mathbf{F}_{S_i}^{RAC} = \text{TE}^{(RAC)} (\mathbf{F}_{S_i}, \mathbf{P}_{S_i}^{RAC}),
\end{equation}
where the positional encoding $\mathbf{P}_{S_i}^{RAC}$ is obtained by conducting a DWConv on $\mathbf{F}$, i.e., $\mathbf{P}^{RAC} = \text{DWConv}(\mathbf{F})$, and masked with $S_i$. $\mathbf{F}_{S_i}^{RAC}$ is the result RAC feature for area $S_i$. Note the embedding operation, which is widely adopted in popular methods \cite{bert,vit}, has been removed. Because the output $\mathbf{F}$ obtained from the backbone network, can be seen as a good feature representation that possesses rich semantic information.

\subsubsection{Global Aggregation Context}

After obtaining RACs, we further build the connections between different areas so that each pixel can perceive global information. The generated contextual information is called GAC, where the ``aggregation'' means the features of these areas are aggregated. Before aggregation, it is necessary to flatten and embed the vectors for each area, and then use a transformer to deal with these embeddings. However, since the shape of these homogeneous areas is various, pixel numbers in these areas are not the same. It is difficult to conduct the embedding for each area with the same group of parameters. Thus, we use a descriptor to represent each area.

In our implementation, the descriptor of each area is obtained by averaging the internal pixel representations, i.e.,
\begin{equation}
  v_i = \frac{1}{n_i} \sum\limits_{j\in S_i} f_j^{RAC},
\end{equation}
where $n_i$ is the pixel number of $S_i$. $f_{j}^{RAC}$ is the pixel representation of $\mathbf{F}_{S_i}^{RAC}$, where position $j \in S_i$.

Then we use a transformer encoder to capture the relationships between these descriptors:
\begin{equation}
  \mathbf{V}^{GAC} = \text{TE}^{(GAC)} (\mathbf{V}, \mathbf{P}^{GAC}),
\end{equation}
where $\mathbf{V}=\{v_1,\cdots,v_Z\} \in \mathbb{R}^{Z \times C}$ is a descriptor set, $\mathbf{P}^{GAC}=\text{DWConv}(\mathbf{V})$. Note that this DWConv adopts 1-D convolution with $1 \times 3$ kernels since $v_i$ is a 1-D vector.

Now each vector in $\mathbf{V}^{GAC}\in \mathbb{R}^{Z \times C}$ has encoded the contexts between different homogeneous areas. It should be further recovered to a 2-D feature map. Thus, we adopt a transformer decoder:
\begin{equation}
  \mathbf{F}^{GAC} = \text{TD}^{(GAC)} (\mathbf{F}^{RAC}, \mathbf{V}^{GAC}).
\end{equation}
Here, $\mathbf{F}^{RAC} = \{\mathbf{F}^{RAC}_{S_1},\cdots,\mathbf{F}^{RAC}_{S_Z}\} \in \mathbb{R}^{C \times H_1 \times W_1} $ is a 2-D feature map, where each area has captured an internal context. Each pixel representation in $\mathbf{F}^{GAC} $ has aggregated the context in $\mathbf{V}^{GAC}$. Since these descriptors represent the homogeneous areas, $\mathbf{F}^{GAC}$ also can be regarded as the extracted GAC feature.

\subsection{Dual Context Network with Transformer}

\begin{figure}[t]
  \centering
  \includegraphics[width=0.8\linewidth]{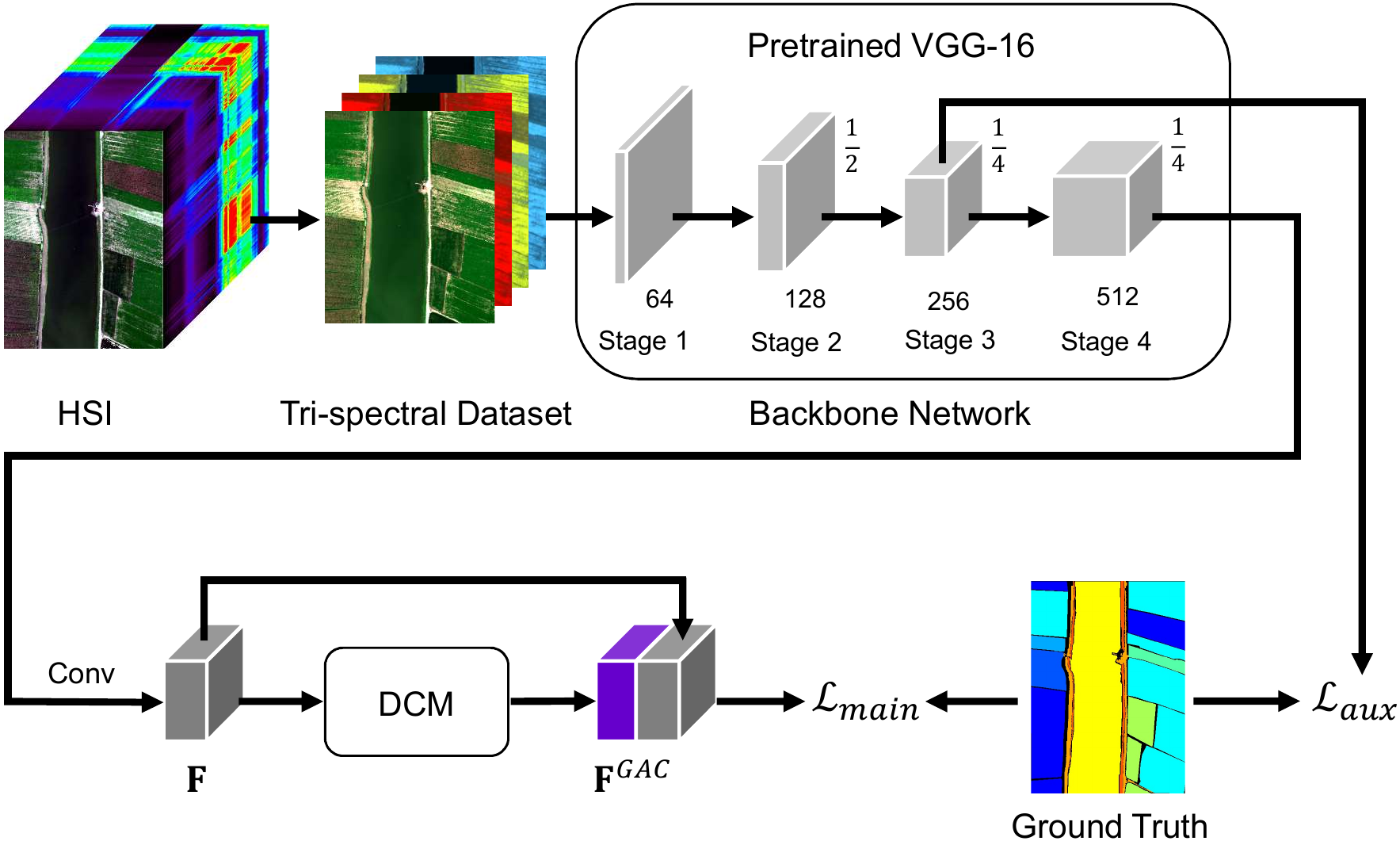}\\
  \caption{The overall architecture of the proposed DCN-T. The generated tri-spectral image set is fed into a modified VGG-16 backbone network. This network adopts the ImageNet pretrained parameters, and an extra added convolutional layer, to produce a high representation feature $\mathbf{F}$ for the subsequent DCM. The output $\mathbf{F}^{GAC}$ of DCM is concatenated with $\mathbf{F}$ for the final prediction. An auxiliary loss is defined on the stage 3 feature in the backbone network.}
  \label{network}
\end{figure}

In this section, we first introduce the architecture of the proposed DCN-T. Since the classification results are from the derived tri-spectral image set, the final prediction can be obtained by voting. The details will be presented later.

\subsubsection{Architecture}

Combining the proposed DCM with an ImageNet pretrained backbone network, we obtain a complete segmentation network named DCN-T, where ``T'' represents the transformer used in DCM. After reviewing the literature in the remote sensing community \cite{wang_rsp_2022}, it has been found that VGG-16 \cite{vgg} is one of the most frequently utilized backbones. As a result, we adopt the VGG-16 network as the primary backbone for our study. We remove the fifth stage and the last three pooling layers to preserve a high-resolution feature map, which has been proven to be effective in segmentation \cite{hrnet}. Therefore, $H_1 = H/4, W_1 = W/4$. We add a 3 $\times$ 3 standard convolution to reduce the dimension to $C=256$ in the feature $\mathbf{F}$. In the DCM, since we use one transformer encoder to obtain RACs, and then use one transformer encoder and a single-layer transformer decoder to separately encode the relationships between homogeneous areas and aggregate global contexts to form 2-D features. Therefore, $N_D = N_E = 1$. In addition, the MLP ratio is always set to 2 for all transformer layers. Finally, the input and output of the DCM, i.e., $\mathbf{F}$ and $\mathbf{F}^{GAC}$, are concatenated together. The prediction is obtained after a convolutional layer and a bilinear upsampling layer. Other hyper-parameters, such as the number of homogeneous areas $Z$ and the head number $h$, will be determined in the ablation experiments.

\noindent \textbf{Loss Function.} Following \cite{pspnet}, except the main loss that is used for the final prediction, we also define an auxiliary loss whose weight equals 0.4, i.e.,
\begin{equation}
  \mathcal{L} = \mathcal{L}_{main} + 0.4\mathcal{L}_{aux}
\end{equation}
The auxiliary loss is set on the output of the third stage in the backbone network. The two losses both adopt the cross entropy loss:
\begin{equation}
  L_{CE} = -\frac{1}{HW}\sum\limits_{i=1}^H \sum\limits_{j=1}^W \sum\limits_{c=1}^{C_n} \widetilde{Y}_{cij} \log(\widehat{P}_{cij}),
\end{equation}
where $\widehat{P} \in \mathbb{R}^{C_n \times H \times W}$ is the probability prediction matrix, which is the output of the upsampling layer and will be used to produce classification map $\widehat{Y}$ by an argmax operation. $C_n$ is the number of classes, $\widetilde{Y} \in \mathbb{R}^{C_n \times H \times W}$ is the one-hot encoding of the labeled map $Y$.

\subsubsection{Voting Prediction}

\begin{figure}[t]
  \centering
  \includegraphics[width=1\linewidth]{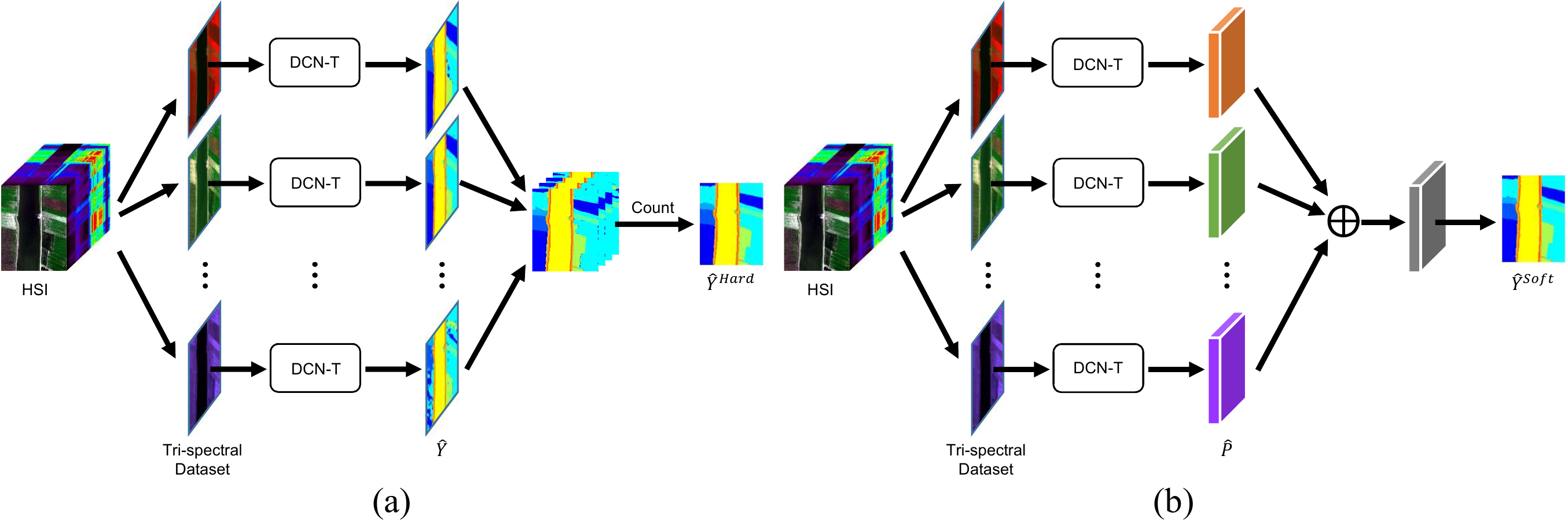}\\
  \caption{Illustration of two voting schemes. (a) Hard Voting. (b) Soft Voting.}
  \label{vote_scheme}
\end{figure}

From each tri-spectral image, we can obtain a classification map. It is necessary to comprehensively utilize different results for boosting accuracy. In this paper, we adopt the voting mechanism, where two schemes are separately explored, i.e., ``hard voting'' and ``soft voting'', see Figure \ref{vote_scheme}.

\noindent\textbf{Hard Voting.} ``Hard voting'' is the most typical voting that follows the principle of ``minority obeys majority''. Assume we have obtained the result maps $\{\widehat{Y}^{(1)},\cdots,\widehat{Y}^{(M)}\}$ from a generated tri-spectral image set that contains $M$ images. Then the voting result for the $i$-th row and $j$-th column pixel is based on the counts of each category: 
\begin{equation}
  \widehat{Y}_{ij}^{Hard} = \mathop{\arg\max}_{c \in \{1,\cdots,C_n\}} \sum\limits_{k=1}^{M} \mathcal{I}(\widehat{Y}_{ij}^{(k)} = c),
\end{equation}
where $\mathcal{I}(\cdot)$ is a binary indicator that judges whether the category of the corresponding pixel in the $k$-th image is $c$.

\noindent\textbf{Soft Voting.} Instead of using off-the-shelf classification maps, the ``soft voting'' considers probabilities that are neglected in ``hard voting'' and combines them by addition:
\begin{equation}
  \widehat{Y}_{ij}^{Soft} = \mathop{\arg\max}_{c \in \{1,\cdots,C_n\}} \sum\limits_{k=1}^{M} \widehat{P}_{ij}^{(k)},
\end{equation}
where $\widehat{P}_{ij}^{(k)}$ is a vector in size of $1 \times C_n$, $\widehat{Y}_{ij}^{(k)} = \mathop{\arg\max}_{c \in \{1,\cdots,C_n\}} \widehat{P}_{ij}^{(k)}$.

\section{Experiment}

\subsection{HSI Dataset}

\begin{table}[t]
  \caption{Important parameters on the three scenes of the WHU-Hi dataset}
  \newcommand{\tabincell}[2]{\begin{tabular}{@{}#1@{}}#2\end{tabular}}
  \centering
  \scriptsize
  \begin{tabular}{|l|c|c|c|}
  \hline
  Scene &Image Size & Number of Channel & Number of Category \\
  \hline
  LongKou & 550 $\times$ 400 &  270 & 9 \\
  \hline
  HanChuan & 1217 $\times$ 303 & 270 & 16 \\
  \hline
  HongHu & 940 $\times$ 475 & 270 & 22  \\
  \hline
\end{tabular}
  \label{dataset_tab}
\end{table}

\begin{figure}[t]
  \centering
  \includegraphics[width=0.8\linewidth]{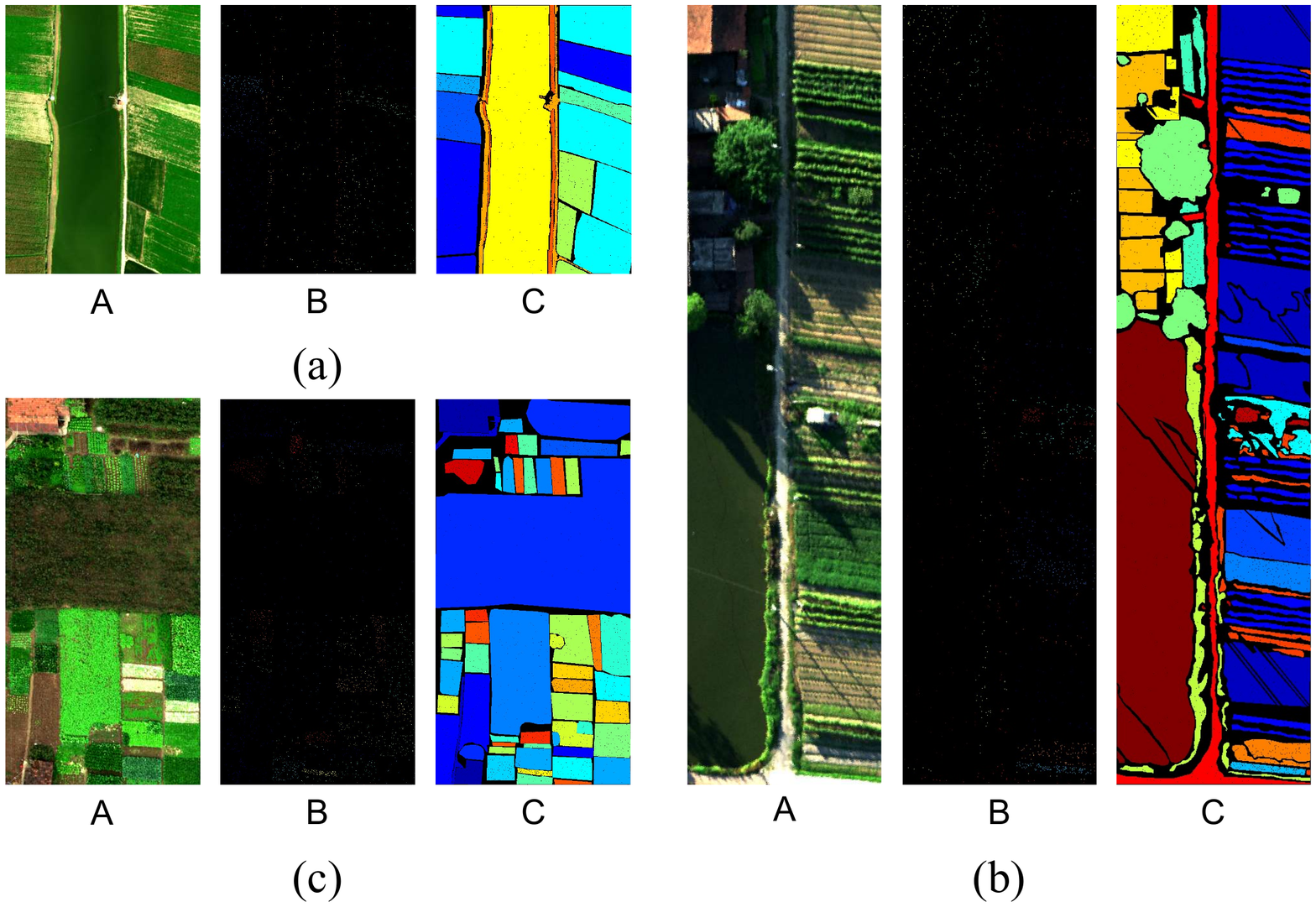}\\
  \caption{The False color image and split ground truth of the WHU-Hi dataset. A, B, and C separately denote the image of the generated tri-spectral dataset, labeled map of the training set (100 samples per class), and labeled map of the testing set. (a) LongKou. (b) HanChuan. (c) HongHu.}
  \label{dataset}
\end{figure}

In this paper, a recent and challenging HSI classification dataset---WHU-Hi dataset is used to evaluate the performance of the proposed method. This dataset possesses high spatial resolution since it is acquired on an unmanned aerial vehicle-borne hyperspectral system. This dataset has three scenes including LongKou, HanChuan, and HongHu, which are obtained in farming areas with diverse crops. The related parameters that are useful for experiments have been listed in Table \ref{dataset_tab}.

Different from previous HSI public datasets such as the Indian Pines or Pavia University in which all samples are knowable, requiring researchers to manually split training and testing sets, causing performances of different methods may be compared unfairly. The training and testing sets of the WHU-Hi dataset have been officially divided. In our research, we use three kinds of ground truth, which have 25, 50, or 100 training labels for each class, respectively. An example including the training set where 100 samples per class are available, and its corresponding testing set (also provided by the official) has been presented in Figure \ref{dataset}.

\subsection{Implementation Details and Experimental Settings}

In the experiments, since the ImageNet pretrained parameters are adopted in our VGG-16 backbone network, the backbone network only needs to be finetuned. For this reason, we separately apply two different learning rates on the backbone network and the proposed DCM to flexibly optimize the whole network. Concretely, the learning rate of the DCM is 10 times of the backbone network. For the backbone network, the initial learning rate is 0.001, which is constantly adjusted with the polynomial scheduling policy: $current\_lr = initial\_lr(1-iter/max\_iter)^{0.9}$. The stochastic gradient descent (SGD) with momentum strategy is used to update the network parameters, where the momentum is set to 0.9 and the weight decay is configured as 0.0001. The training epoch is set to 30 for all scenes. Considering the image size of these scenes, the batch sizes of LongKou, HanChuan, and HongHu are separately adopted to 4, 3, and 2. Three commonly used metrics in the HSI classification community including the overall accuracy (OA), average accuracy (AA), and kappa coefficient ($\kappa$) are employed to evaluate the performance of the proposed method. The method is implemented with the Pytorch framework and conducted on a single 16G NVIDIA Tesla V100 GPU.

Experiments are organized as follows. We first conduct a series of hyper-parameter analyses to find an optimal model configuration. Then, ablation studies are conducted to separately demonstrate the effectiveness of the proposed tri-spectral image generation pipeline, the presented DCM, and the finally obtained DCN-T. In addition, performance comparisons between the proposed method and existing classical or state-of-the-art approaches including some recent transformer-based works will also be depicted in the subsequent section. Besides quantitative experiments, some qualitative investigations, including the exhibition of the generated tri-spectral dataset and visualization of internal features, make the proposed method more convincible.

\subsection{Parameter Analysis}

Several hyperparameters are more suitable to be determined through experiments rather than experiences. Since they tend to be involved in the characteristics of data itself or the core components of the network, accordingly affecting the model performance.

\begin{table}[t]
  \caption{Influence of the hyper-parameter $G$ in the proposed method on the LongKou Scene in the WHU-Hi dataset (100 training samples per class, hard voting, unit of OA: \%, unit of times: s). The time is counted on a single A100 GPU}
  \newcommand{\tabincell}[2]{\begin{tabular}{@{}#1@{}}#2\end{tabular}}
  \centering
  \scriptsize
  \begin{tabular}{|l|c|c|c|c|c|c|c|}
  \hline
  $G$ & 3 & 5 & 6 & 9 & 10 & 15 & 18 \\
  \hline\hline
  $M$ & 1 & 10 & 20 & 84 & 120 & 455 & 816 \\
  \hline
  OA & 94.72 &  97.47 & 97.69 & 98.32 & 98.57 & 98.78 & 98.67 \\
  \hline
  $T_{trn}$ & 62.70 &99.87  & 226.66  &  927.66 & 1302.84 & 5248.48 & 8472.32 \\
  \hline
  $T_{tes}$ & 5.92 & 7.63 & 9.15 & 21.08 & 27.15  & 80.02  & 149.20 \\
  \hline
\end{tabular}
  \label{group_number}
\end{table}

\noindent\textbf{Channel Group Number} $G$. When generating a tri-spectral dataset, we need to reduce dimension while exploiting the abundant spectral information as much as possible (In fact, the spectral information is transferred to different tri-spectral images, which will be shown in later visualization). Our target also includes finding an optimal scale to produce tri-spectral channels. Since $G$ determines the capacity $M$ of the generated tri-spectral dataset, it also has an influence on the training and testing time (denoted as $T_{trn}$ and $T_{tes}$). Thus, we first investigate the influence of $G$. Since the number of channels in the WHU-Hi dataset is 270. $G$ is limited as a divisor of 270 to leverage all channels. Therefore, the smallest value of $G$ is 3 (at this time $M=1$, and the corresponding epoch is set to 100, while the epoch for other $G$ is still 30). Table \ref{group_number} indicates that the smallest $G$ leads to the worst performance since many useful spectral details are discarded through directly averaging a large number of channels. The accuracy of the proposed method will be increased when enlarging $G$ since more spectral information can be kept. Accordingly, the required training and testing time are also increased, because the capacity of the tri-spectral dataset is expanded. However, if we further increase $G$ from 15 to 18, although more tri-spectral images are produced, the accuracy is instead decreased. We think that a large $G$ causes the curse of dimensionality that can not be effectively eliminated by subsequent aggregation. If adjacent groups are similar, it may produce some relatively low-quality tri-spectral images that are not beneficial to classification because different channels are correlated. Therefore, in our practice, $G$ is set to 15. Correspondingly, $M=455$ tri-spectral images are used for voting.

\begin{table}[t]
  \caption{The OAs (\%) with different $Z$ of the proposed method on the LongKou Scene in the WHU-Hi dataset (100 training samples per class, hard voting)}
  \newcommand{\tabincell}[2]{\begin{tabular}{@{}#1@{}}#2\end{tabular}}
  \centering
  \scriptsize
  \begin{tabular}{|l|c|c|c|c|c|}
  \hline
  $Z$ & 16 & 32 & 64 & 128 & 256 \\
  \hline\hline
  OA & 98.64 & 98.67 &98.73 & 98.78 & 98.70 \\
  \hline
\end{tabular}
  \label{area_number}
\end{table}

\noindent\textbf{Homogeneous Area Number} $Z$. Since RAC and GAC are captured on the foundation of the generated homogeneous areas. The setting of these areas is important. It is easy to observe that the number of homogeneous areas is also the number of descriptors that are used for global context perception. Thus, $Z$ depicts the recapitulation degree of these descriptors to represent the whole scene. Too few descriptors are more abstract and cannot effectively represent the contents, because each descriptor is obtained from a large area that may simultaneously contain multiple different objects, affecting the performance of the obtained GAC. While a large $Z$ not only causes redundancy but also increases computational complexity, and has a risk of overfitting. In addition, $Z$ is also inversely proportional to the average size of these homogeneous areas. Thus, $Z$ also determines the range of RAC capturing. We test different $Z$ to achieve a trade-off between global and local. $Z$ is restricted as an integer multiple of 2. $Z$ is from 16 to 256, and the results are shown in Table \ref{area_number}. It can be seen that, in the early stage, with the increase of $Z$, accuracies are constantly improved. However, the model performances are affected when generating too many homogeneous areas, which can be exemplified by $Z=256$. Thus, in later ablation studies and performance comparisons, $Z$ will be set to 128 to generate suitable homogeneous areas for subsequent DCM.

\begin{table}[t]
  \caption{The GPU memory usages (MB) with different $h$ on three scenes in the WHU-Hi dataset}
  \newcommand{\tabincell}[2]{\begin{tabular}{@{}#1@{}}#2\end{tabular}}
  \centering
  \scriptsize
  \begin{tabular}{|l|c|c|c|c|}
  \hline
  $h$ & 1 & 2 & 4 & 8 \\
  \hline\hline
  LongKou (Batchsize=4)&  10322    &  10456     &  10642    & 11106 \\
  \hline
  HanChuan (Batchsize=3)& 11908 & 12116 & 12524 & 13344 \\
  \hline
  HongHu (Batchsize=2) & 10704 & 10960 & 11354 & 12100 \\
  \hline
\end{tabular}
  \label{gpu_h}
\end{table}

\begin{table}[t]
  \caption{The OAs (\%) with different $h$ of the proposed method on the LongKou Scene in the WHU-Hi dataset (100 training samples per class, hard voting)}
  \newcommand{\tabincell}[2]{\begin{tabular}{@{}#1@{}}#2\end{tabular}}
  \centering
  \scriptsize
  \begin{tabular}{|l|c|c|c|c|}
  \hline
  $h$ & 1 & 2 & 4 & 8 \\
  \hline\hline
  OA & 98.77 & 98.77 &98.78 & 98.78 \\
  \hline
\end{tabular}
  \label{acc_h}
\end{table}

\noindent\textbf{Head Number $h$}. Since the particularity of the transformer is mainly from MHSA and MHA, it is necessary to conduct an exploration for a more powerful context capturing. To this end, we separately evaluate the network with different $h$. For simplicity, the $h$s in the transformer encoder and decoder are set to the same and are divisible by the number of channels in the transformer input. Theoretically, the representation ability of the transformer is boosted with the improvement of $h$. Nevertheless, model complexity is naturally increased, causing a large GPU memory usage and even out of memory, especially for more complicated scenes such as the HanChuan and HongHu (see Table \ref{gpu_h}). Therefore, we only test four situations where $h$ is separately configured to 1, 2, 4, and 8. However, Table \ref{acc_h} shows that although relatively large $h$s correspond to higher accuracies, accuracies change little. Thus, we adopt $h=4$ in later experiments.

\subsection{Ablation Study}

We separately conduct a series of ablation studies to explore or show the effectiveness of the proposed method.

\begin{table}[t]
  \caption{The OAs (\%) of the proposed tri-spectral image generation pipeline and different variants on the LongKou Scene in the WHU-Hi dataset (100 training samples per class, hard voting)}
  \newcommand{\tabincell}[2]{\begin{tabular}{@{}#1@{}}#2\end{tabular}}
  \centering
  \scriptsize
  \begin{tabular}{|l|c|c|c|c|}
  \hline
  Pipeline & OA \\
  \hline\hline
  All in & 97.48 \\
  \hline
  Mapping to tri-spectral & 93.75  \\
  \hline
  Neighbored (w/o stretching) & 97.08\\
  Neighbored (w stretching) & 97.46\\
  \hline
  Proposed (w/o stretching) & 98.55 \\
  Proposed (w channel-by-channel stretching) &  97.92 \\
  Proposed (w stretching) &  98.78 \\
  \hline
\end{tabular}
  \label{rgb_variants}
\end{table}

\noindent\textbf{Tri-spectral Generation Pipeline.}  We try diverse variants to evaluate the designed pipeline. The first variant that we assess is an ``All in'' strategy, where the HSI including all channels is directly fed into the network. Naturally, the input channel number of the first convolutional layer is correspondingly changed. This strategy has been adopted in previous image-level approaches, such as \cite{ssfcn}. Table \ref{rgb_variants} shows that this setting is not great, since the channels that have not been refined are used at once, where corresponding issues still exist. We also attempt to add a 1 $\times$ 1 convolution as a mapping layer before the network for directly projecting the HSI to a tri-spectral image. However, this setting performs the worst, because a lot of useful spectral information is unavoidably lost. We adopt a ``neighbored'' strategy, where adjacent three channels in the original HSI are directly employed to generate the tri-spectral image and verify the influence of complete spectral information. The stride in the ``neighbored'' strategy equals 1 with no padding and we obtain a total of 268 tri-spectral images. Corresponding results indicate that the performance of this strategy is even not as good as the ``all in'' strategy because the richness of the obtained tri-spectral dataset is limited. 

We also evaluate the impact of ``stretching'', and the results have been presented in Table \ref{rgb_variants}. Since stretching makes the generated tri-spectral image closer to the natural image that has a high contrast ratio. It is more conducive to applying the ImageNet pretrained parameters. We also test a channel-by-channel stretching, however, the accuracy is decreased. In our consideration, spectral information is indicated by the relative values between different bands, while the channel-by-channel stretching breaks these relationships. Compared to the above variants, our designed pipeline that simultaneously adopts the ``grouping and aggregation'' and ``stretching'' strategy achieves the best, since the spectral information has been fully exploited. The high-dimensional related issues are effectively alleviated, generating high-quality tri-spectral images.

\begin{figure}[t]
  \centering
  \includegraphics[width=0.95\linewidth]{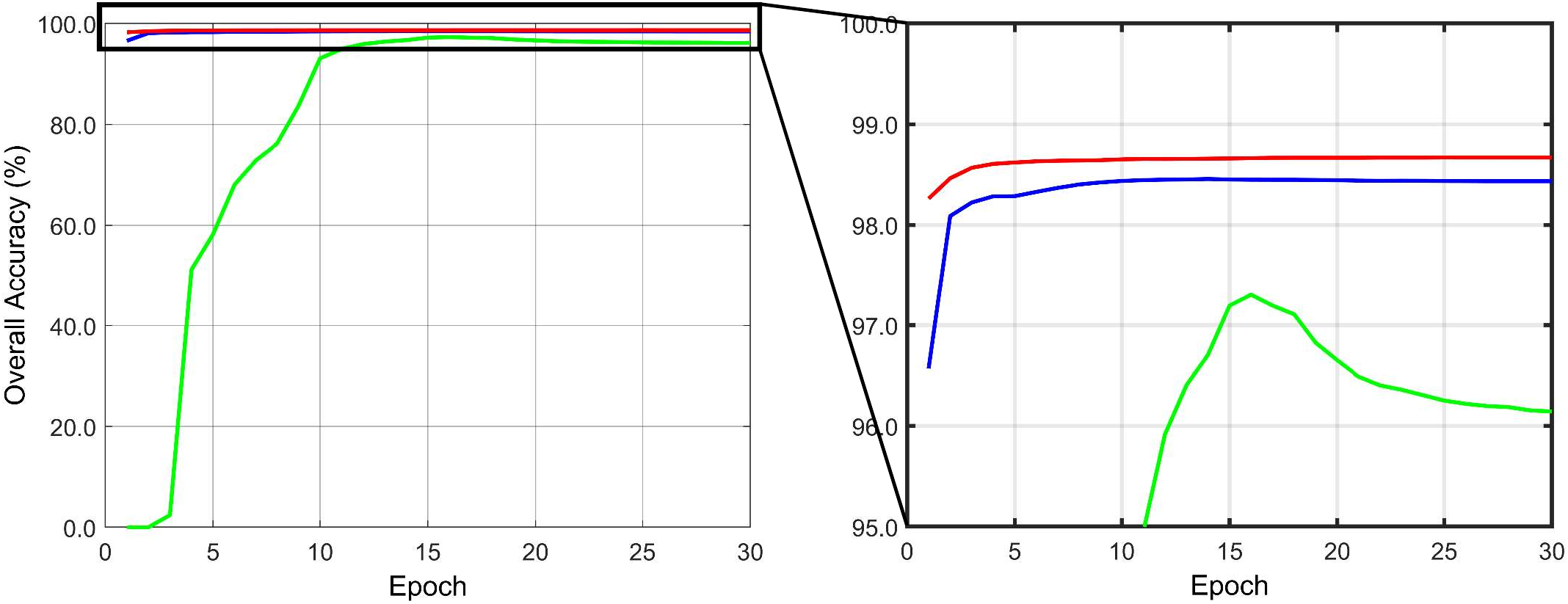}\\
  \caption{The OA changes on the validation set under three different settings in the training stage (100 training samples per class, hard voting). Red line: Using the tri-spectral image with the ImageNet pretrained parameters (the proposed method). Blue line: Using the tri-spectral image without the ImageNet pretrained parameters. Green line: The ``All in'' strategy without the ImageNet pretrained parameters}
  \label{pretrained}
\end{figure}

\noindent\textbf{Imagenet Pretrained Parameters.}
The reason why we use the tri-spectral image is that we would like to address the spatial variability of the HSI by the ImageNet pretrained parameters, which have strong representative abilities. We present different OA changed curves in Figure \ref{pretrained} to show our motivation. It can be observed that, compared with the blue curve, with the pretrained parameters, higher initial accuracies can be obtained and the network converges faster (see red line). In addition, a comparison between the blue line and green line indicates the necessity of using tri-spectral images. Note that the OA in Figure \ref{pretrained} is evaluated on the validation set, which uses 5\% images split from the training set and adopts the same ground truth. The evaluated accuracies are also obtained by voting.

\begin{table}[t]
  \caption{The OAs (\%) with different settings of the proposed context capturing modules on the LongKou Scene in the WHU-Hi dataset (100 training samples per class, hard voting)}
  \newcommand{\tabincell}[2]{\begin{tabular}{@{}#1@{}}#2\end{tabular}}
  \centering
  \scriptsize
  \begin{tabular}{|l|ccc|c|}
  \hline
  Network & $\mathbf{F}$ & RAC &  GAC & OA \\
  \hline\hline
  FCN & \ding{51} &  &  & 98.41 \\
  \hline
  RCN-T &  & \ding{51} &   &  98.55\\
  RCN-T & \ding{51} &\ding{51} &   & 98.74\\
  \hline
  GCN-T &  &   & \ding{51}  & 98.44 \\
  GCN-T & \ding{51} & & \ding{51}  & 98.62 \\
  \hline
  DCN-T &   &\ding{51} &\ding{51} & 98.77 \\
  DCN-T & \ding{51} &\ding{51} &\ding{51} & 98.78 \\
  \hline
\end{tabular}
\label{ablation_dcm}
\end{table}

\begin{figure*}[t]
  \centering
  \includegraphics[width=0.6\linewidth]{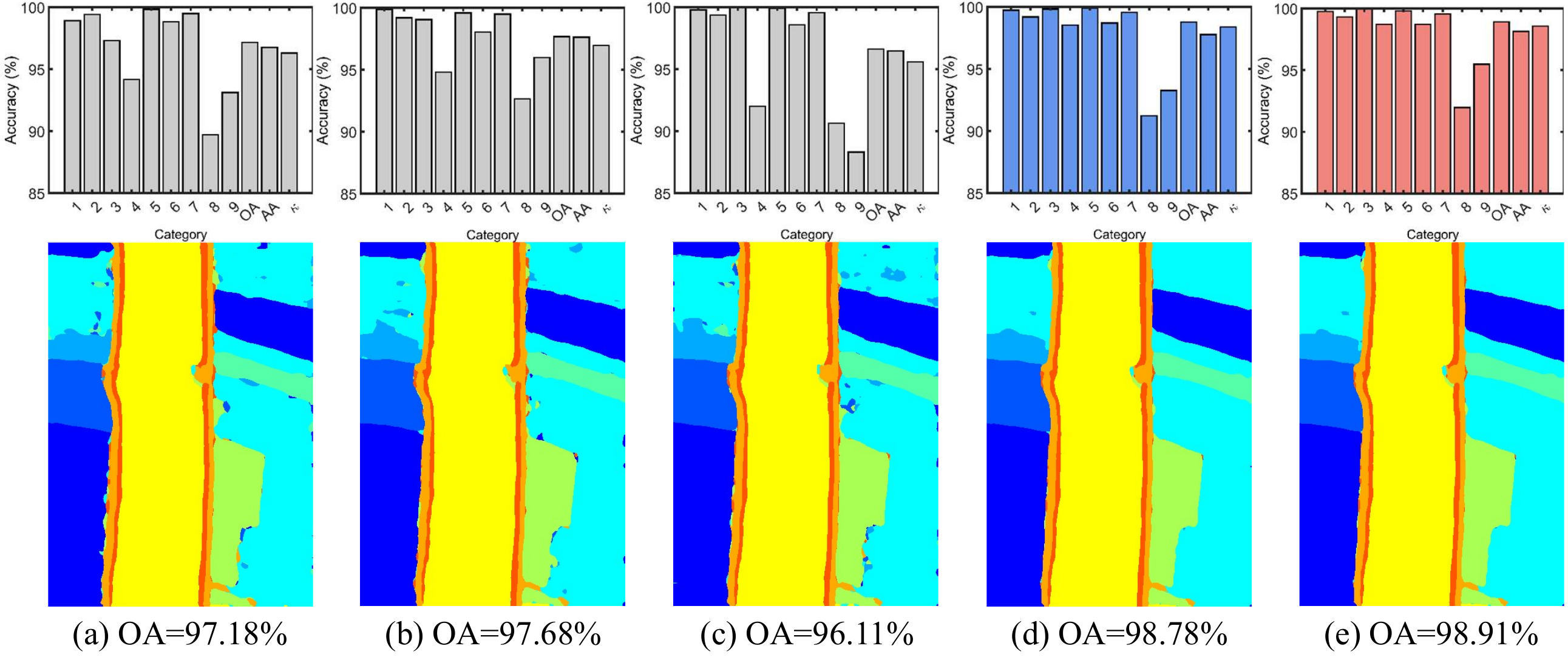}\\
  \caption{The impacts of voting. (a), (b) and (c) are the accuracy of each class (we also include the OA, AA, and $\kappa$ for comparison) and the classification maps. They are obtained from three different tri-spectral images. (d) and (e) are qualitative and quantitative results of hard or soft voting using all 455 images on the LongKou Scene of the WHU-Hi dataset (100 training samples per class, best viewed in color and zoom-in).}
  \label{vote_result}
\end{figure*}

\noindent\textbf{Dual Context Module.}
 We separately adopt various combinations of $\mathbf{F}^{RAC}, \mathbf{F}^{GAC}$ and feature $\mathbf{F}$ to evaluate the proposed DCM. Note that the evaluations of these combinations all require extra training a new network, and the results have been shown in Table \ref{ablation_dcm}. Here, the ``FCN'' directly upsamples $\mathbf{F}$ for classification. ``RCN-T'' and ``GCN-T'' separately denote only RAC or GAC are captured to distinguish from the proposed DCN-T. Table \ref{ablation_dcm} shows that the obtained RAC and GAC both can help classification, and the RAC is more useful than GAC. If we don't conduct intra-area context perception in advance, the improvement that inter-area context capturing brings is limited, because regional details are ignored. In addition, we can see that the accuracies of our DCN-T improve little when considering $\mathbf{F}$. In our thinking, it may be because the valuable information lying in $\mathbf{F}$ has almost been fully leveraged by the designed context modules (in the LongKou scene). The above results show that the obtained features whose pixels can perceive the global by separately capturing the RAC and GAC have powerful representation capabilities, proving the effectiveness of the proposed method.

 \begin{table}[t]
  \caption{The OAs (\%) of conducting the clustering on feature $\mathbf{F}$ or input image on the LongKou Scene in the WHU-Hi dataset (100 training samples per class, hard voting)}
  \newcommand{\tabincell}[2]{\begin{tabular}{@{}#1@{}}#2\end{tabular}}
  \centering
  \scriptsize
  \begin{tabular}{|l|c|c|}
  \hline
  Location & Feature $\mathbf{F}$ & Image \\
  \hline\hline
  OA & 98.78 & 98.60 \\

  \hline
\end{tabular}
  \label{clustering}
\end{table}

\noindent\textbf{Clustering.} The generated homogeneous areas have an influence on the regional context modeling and the outputs of subsequent operations, such as descriptors and recovered global aggregation context features. Usually, compared with internal features of the network, previously obtained tri-spectral images preserve more details and have clearer contours. In view of this, we investigate the performance of clustering on feature $\mathbf{F}$ or input image. Note that the obtained homogeneous area map needs to be downsampled to match the size of feature $\mathbf{F}$, while other steps keep the same. As shown in Table \ref{clustering}, clustering on feature $\mathbf{F}$ performs slightly better. In our consideration, after a series of operations of the CNN such as convolutions and poolings, the object contours of feature $\mathbf{F}$ cannot match with the input image. In addition, clustering on the input image always produces fixed homogeneous areas, which cannot be adaptively adjusted by the changed $\mathbf{F}$ during the training. Therefore, it is better to implement the clustering on the feature $\mathbf{F}$.

\noindent\textbf{Voting.} The OAs in the above parameter analyses and ablation studies have not changed much because results are recorded after voting, which may remedy the classes that have poor accuracies. Since we have generated a tri-spectral dataset, where each image can produce a classification map or probability matrix. We adopt hard voting and soft voting schemes, where the results are merged from 455 images, and the effectiveness has been shown in Figure \ref{vote_result}. It can be seen that the misclassifications and noises have been successfully removed, and almost all classes are improved. In addition, we think voting is a full utilization of spectral information, which is significant for HSI.

\subsection{Performance Comparison}

\begin{table*}[t]
  \caption{The accuracies (\%) of different methods on three scenes in the WHU-Hi dataset (100 training samples per class, HV: Hard Voting, SV: Soft Voting)}
  \newcommand{\tabincell}[2]{\begin{tabular}{@{}#1@{}}#2\end{tabular}}
  \centering
  \resizebox{1\linewidth}{!}{
  \begin{tabular}{|l|l|cccc  c ccc c c c cc|}
  \hline
  \multirow{4}{*}{Scene}&Type & \multicolumn{5}{c|}{Non Transformer} &  \multicolumn{8}{c|}{Transformer-Based}\\
  \cline{2-15}
  & Input & \multicolumn{2}{c|}{Pixel or Patch} & \multicolumn{3}{c|}{Whole Image}  & \multicolumn{5}{c|}{Pixel or Patch} &  \multicolumn{3}{c|}{Whole Image} \\
  \cline{2-15}
  & Method & 3DCNN & SSAN &  SSFCN & ENL-FCN & WFCG & HSI-BERT & SpecFormer & T-SST & SSFTT & LSFAT & LESSFormer & Ours (HV) & Ours (SV)\\
  \hline
  \multirow{3}{*}{LongKou} & OA & 98.26 & 96.81 &  97.20 & 98.05 & 98.70 & 98.49 & 93.75 & 91.12 & 98.52 & 98.67 & 96.41 & \textcolor{blue}{98.78} & \textbf{98.91} \\
  & AA & \textcolor{blue}{98.36} & 96.99 & 97.42 & 96.81 & 98.91 & \textbf{99.05} & 91.98 & 91.22 & 98.70 & 98.75 & 95.23 & 97.77 & 98.14 \\
  & $\kappa$ & 97.72 & 95.83 & 97.44 & 97.54 & 98.30 & 98.03 & 91.89 & 88.54 & 98.06 & 98.25 & 95.32& \textcolor{blue}{98.39} & \textbf{98.57} \\
  \hline
  \multirow{3}{*}{HanChuan} & OA & 94.66 & 71.27 & 73.75 & 86.15 & 92.34 & 94.95 & 75.35 & 84.86 & 94.45 & 95.14 & 87.03 & \textcolor{blue}{96.27} & \textbf{96.38} \\
  & AA & 94.72 & 71.07 & 72.53 & 80.67 & 89.19 & 94.69 & 76.86 & 83.21 & 94.53& 95.24 & 83.68 & \textcolor{blue}{96.83} & \textbf{96.92} \\
  & $\kappa$ & 93.77 & 66.94 & 69.91 & 83.85 & 91.03 & 94.10 & 71.92 & 82.14 &93.52 &94.33 & 84.86 & \textcolor{blue}{95.64} & \textbf{95.77} \\
  \hline
  \multirow{3}{*}{HongHu} & OA & 94.83 & 87.96 & 88.01  & 81.22 & 93.65 & 95.63 & 79.03 & 51.07 & 93.75 & 94.23 & 80.01 & \textcolor{blue}{95.69} & \textbf{95.85} \\
  & AA & 95.14 & 85.91 &  85.95 & 72.75 & 89.70 & 96.06 & 75.87 & 41.08 & 96.05 & 95.02 & 80.83 & \textcolor{blue}{96.64} & \textbf{96.80} \\
  & $\kappa$ & 93.50 & 84.84 &  85.00 & 76.30 & 91.96 & 94.50 & 74.32 & 41.85 & 92.20 & 92.75 & 75.44& \textcolor{blue}{94.56} & \textbf{94.76} \\
  \hline
\end{tabular}
  }
  \label{performance_compare}
\end{table*}

\begin{figure}[t]
  \centering
  \subfigure[]{\includegraphics[width=0.13\linewidth]{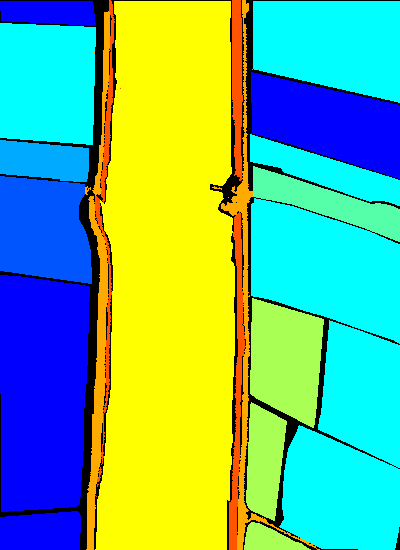}}
  \subfigure[]{\includegraphics[width=0.13\linewidth]{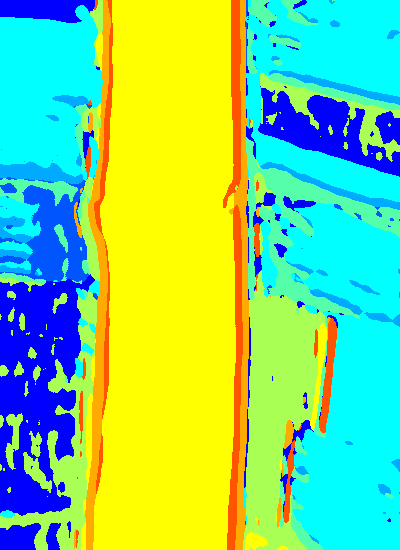}}
  \subfigure[]{\includegraphics[width=0.13\linewidth]{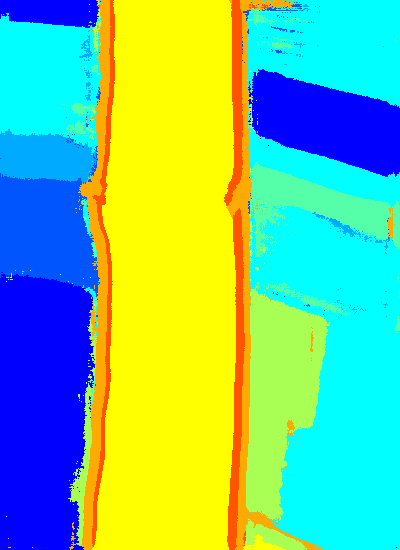}}
  \subfigure[]{\includegraphics[width=0.13\linewidth]{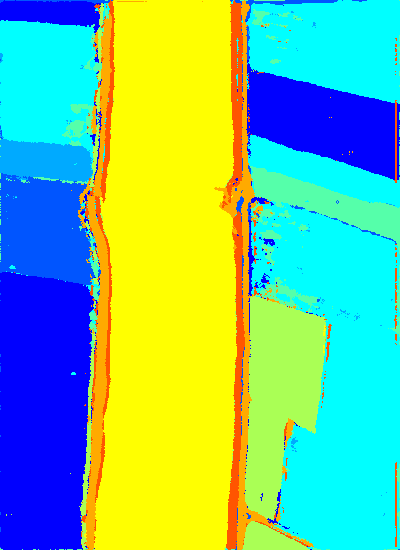}}
  \subfigure[]{\includegraphics[width=0.13\linewidth]{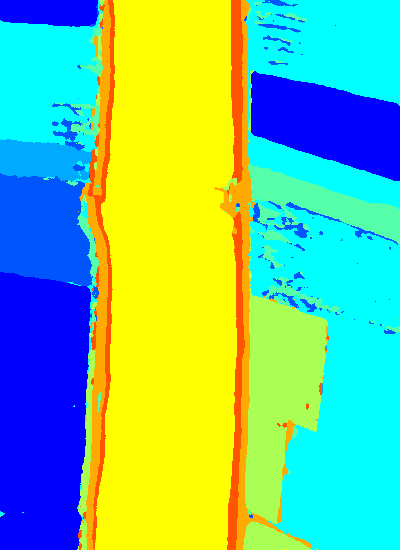}}
  \subfigure[]{\includegraphics[width=0.13\linewidth]{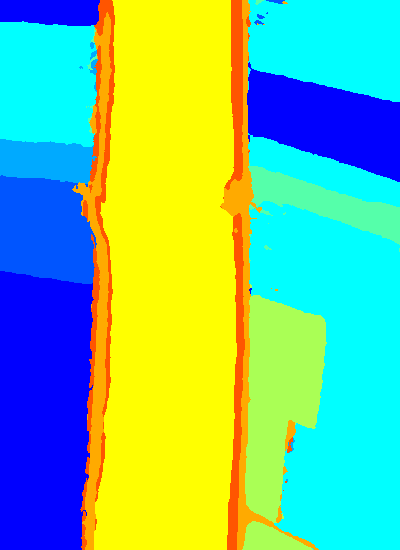}}
  \subfigure[]{\includegraphics[width=0.13\linewidth]{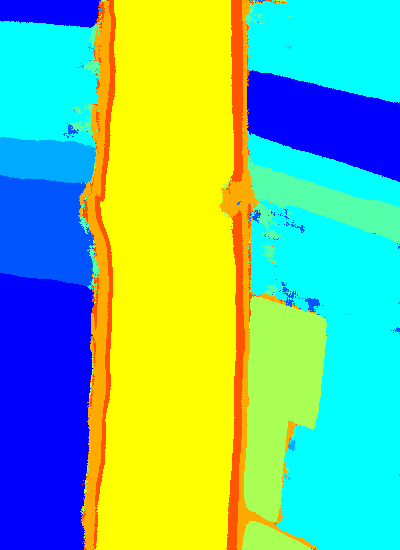}}\\
  \subfigure[]{\includegraphics[width=0.13\linewidth]{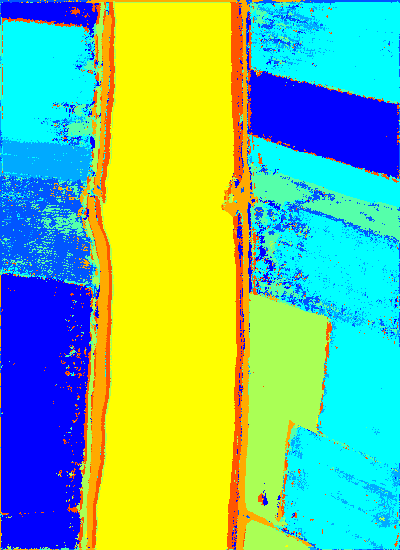}}
  \subfigure[]{\includegraphics[width=0.13\linewidth]{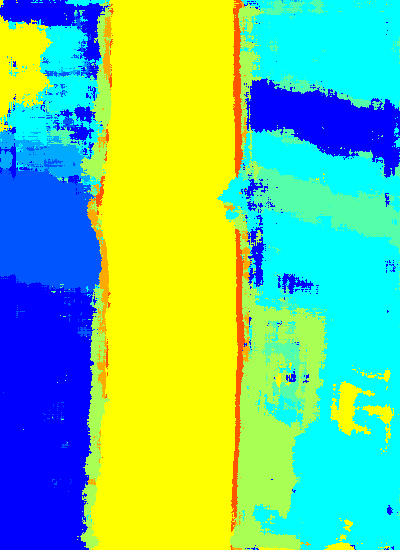}}
  \subfigure[]{\includegraphics[width=0.13\linewidth]{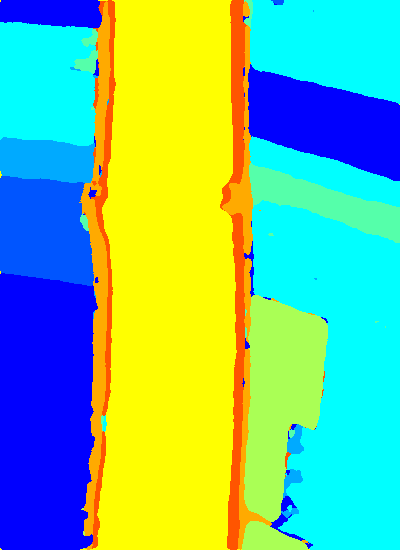}}
  \subfigure[]{\includegraphics[width=0.13\linewidth]{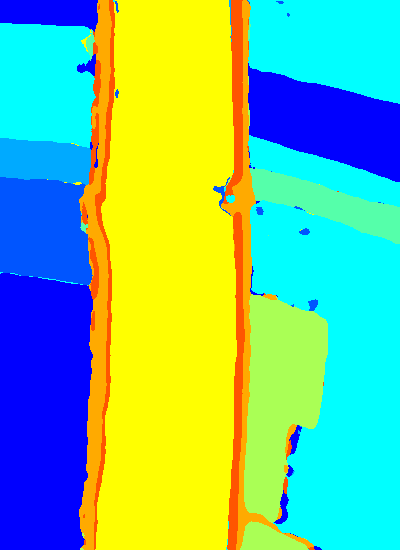}}
  \subfigure[]{\includegraphics[width=0.13\linewidth]{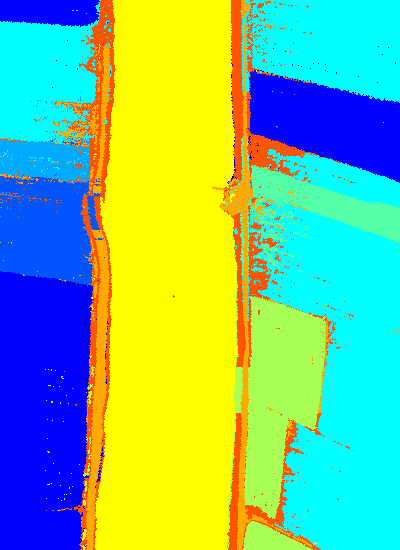}}
  \subfigure[]{\includegraphics[width=0.13\linewidth]{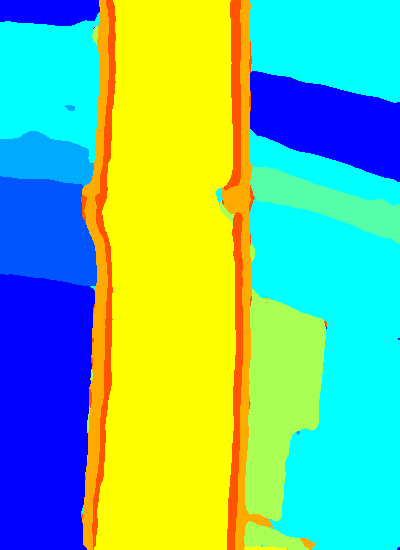}}
  \subfigure[]{\includegraphics[width=0.13\linewidth]{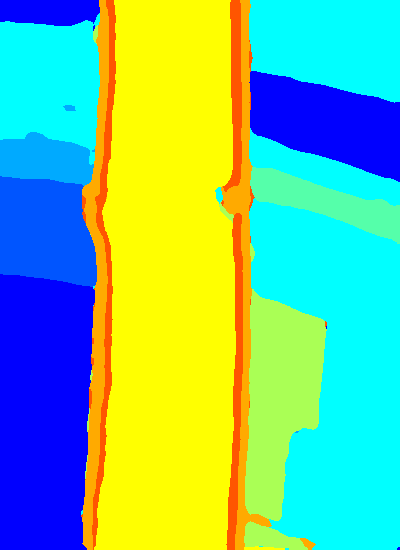}}
  \caption{Classification maps of different methods on the LongKou Scene in the WHU-Hi dataset (100 training samples per class). (a) Ground Truth. (b) 3DCNN. (c) SSAN. (d) SSFCN. (e) ENL-FCN. (f) WFCG. (g) HSI-BERT. (h) SpecFormer. (i) T-SST. (j) SSFTT. (k) LSFAT. (l) LESSFormer. (m) Ours (Hard Voting). (n) Ours (Soft Voting).}
  \label{longkou_cls}
\end{figure}

\begin{figure}[t]
  \centering
  \subfigure[]{\includegraphics[width=0.08\linewidth]{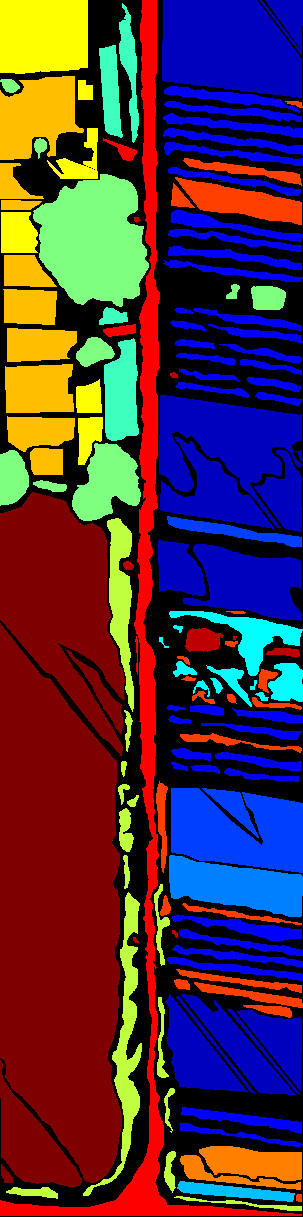}}
  \subfigure[]{\includegraphics[width=0.08\linewidth]{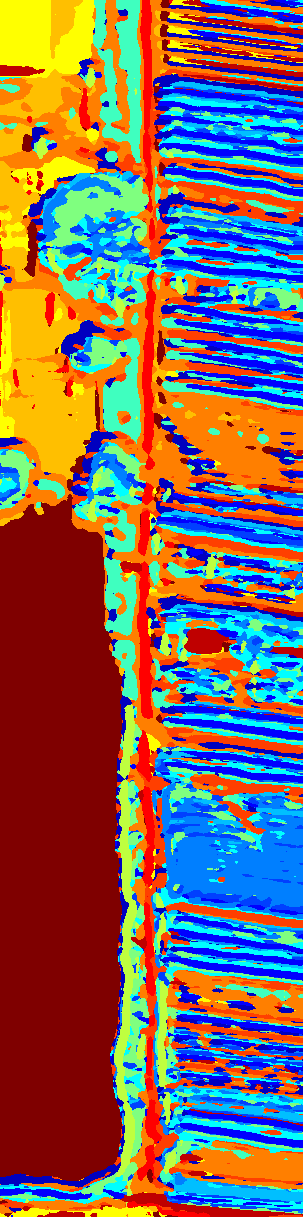}}
  \subfigure[]{\includegraphics[width=0.08\linewidth]{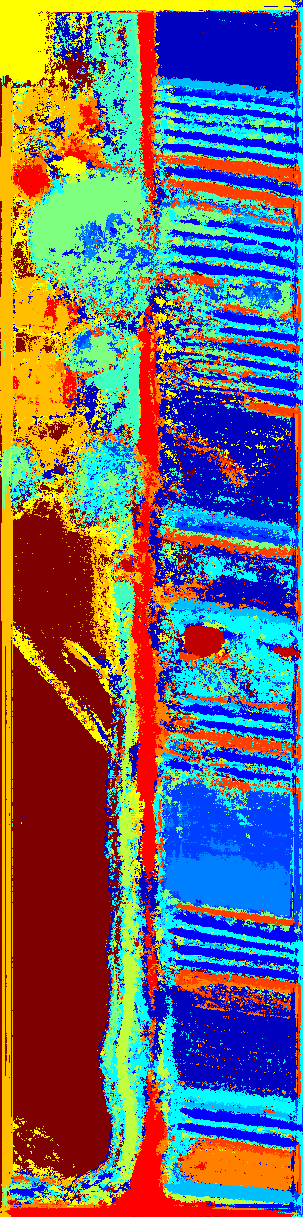}}
  \subfigure[]{\includegraphics[width=0.08\linewidth]{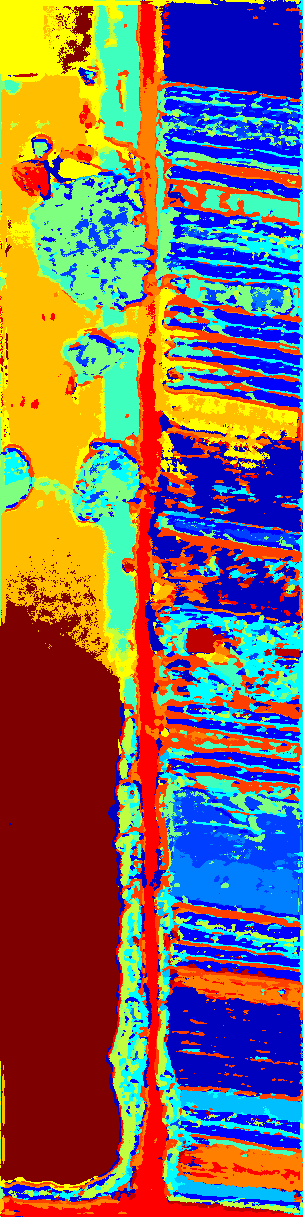}}
  \subfigure[]{\includegraphics[width=0.08\linewidth]{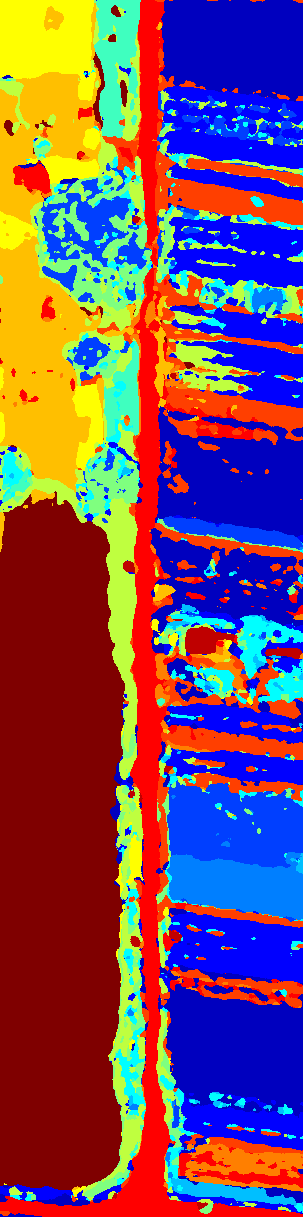}}
  \subfigure[]{\includegraphics[width=0.08\linewidth]{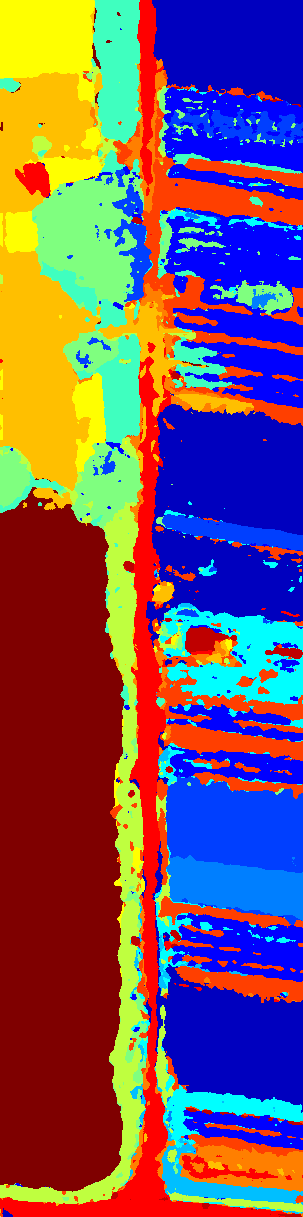}}
  \subfigure[]{\includegraphics[width=0.08\linewidth]{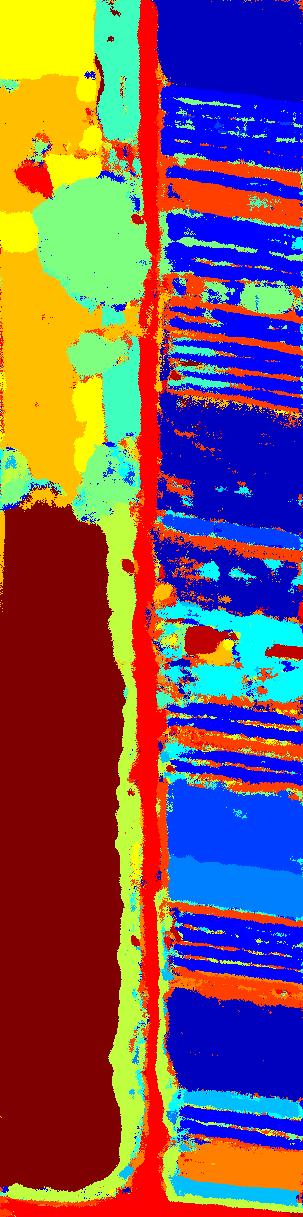}}\\
  \subfigure[]{\includegraphics[width=0.08\linewidth]{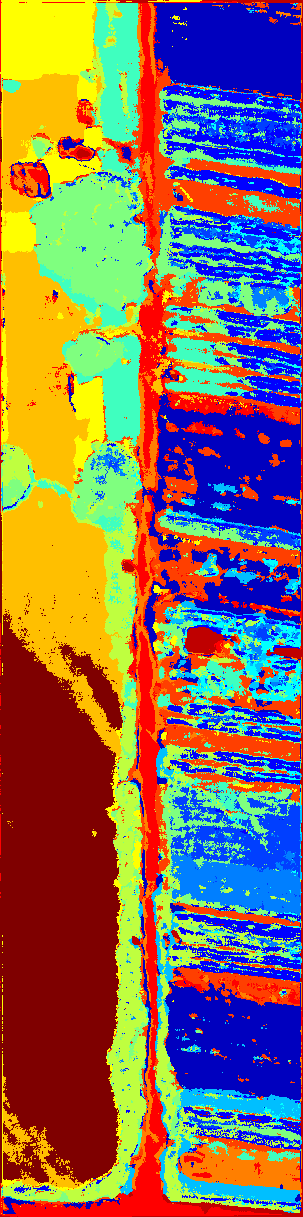}}
  \subfigure[]{\includegraphics[width=0.08\linewidth]{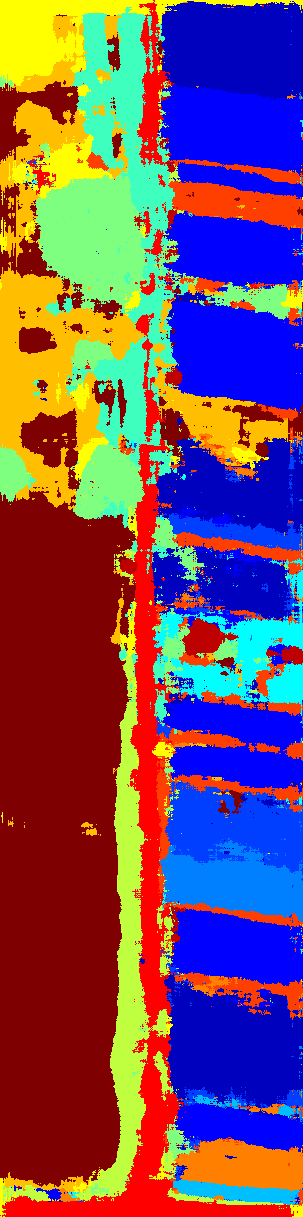}}
  \subfigure[]{\includegraphics[width=0.08\linewidth]{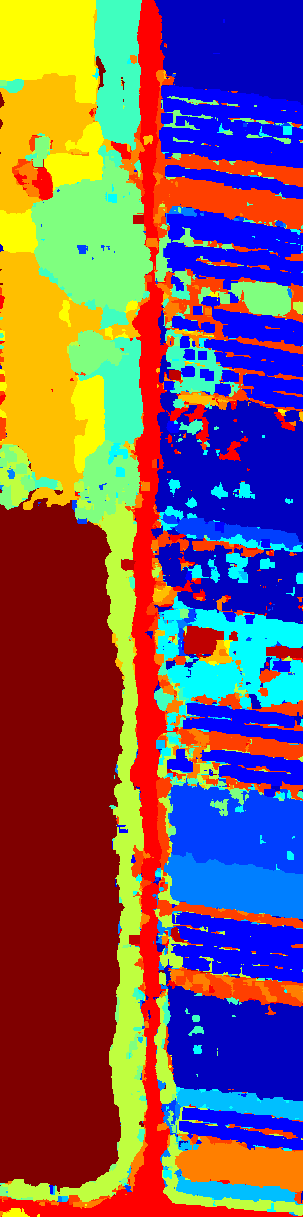}}
  \subfigure[]{\includegraphics[width=0.08\linewidth]{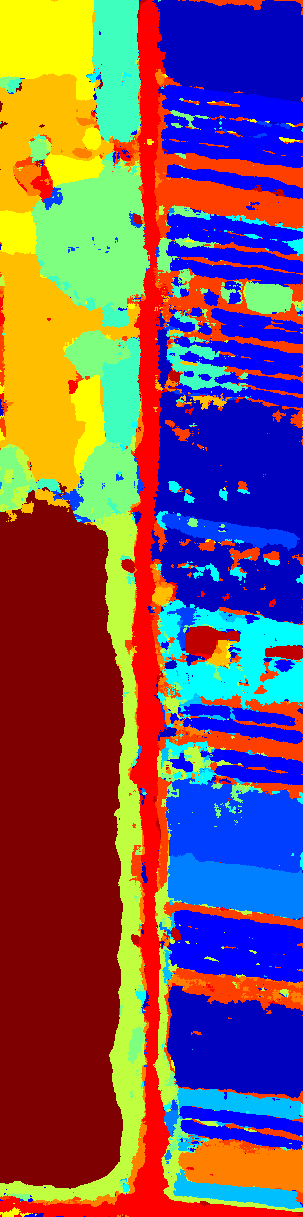}}
  \subfigure[]{\includegraphics[width=0.08\linewidth]{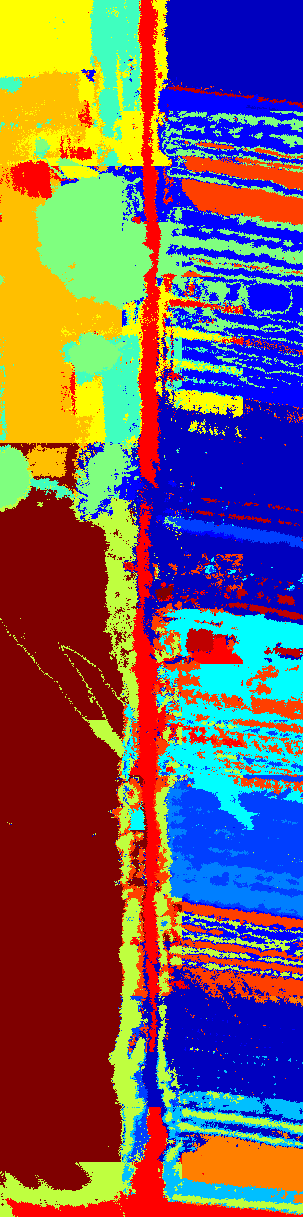}}
  \subfigure[]{\includegraphics[width=0.08\linewidth]{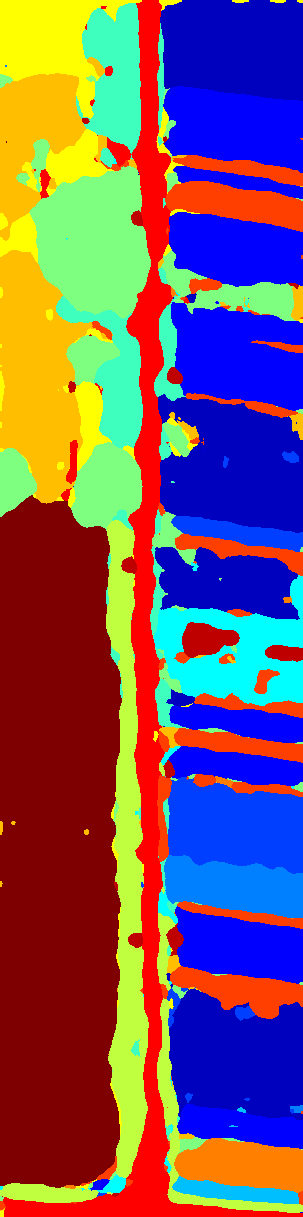}}
  \subfigure[]{\includegraphics[width=0.08\linewidth]{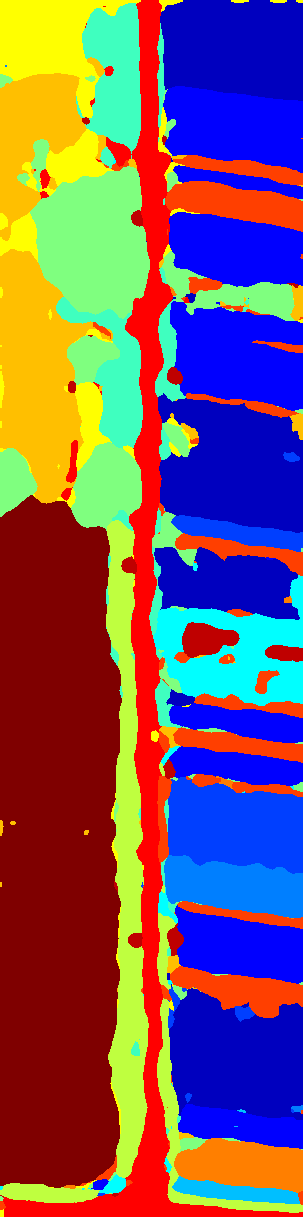}}
  \caption{Classification maps of different methods on the HanChuan Scene in the WHU-Hi dataset (100 training samples per class). (a) Ground Truth. (b) 3DCNN. (c) SSAN. (d) SSFCN. (e) ENL-FCN. (f) WFCG. (g) HSI-BERT. (h) SpecFormer. (i) T-SST. (j) SSFTT. (k) LSFAT. (l) LESSFormer. (m) Ours (Hard Voting). (n) Ours (Soft Voting).}
  \label{hanchuan_cls}
\end{figure}

\begin{figure}[t]
  \centering
  \subfigure[]{\includegraphics[width=0.13\linewidth]{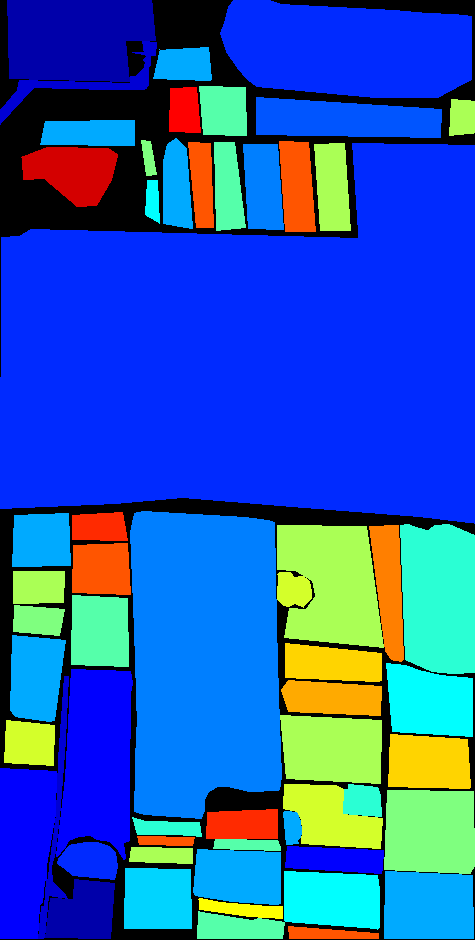}}
  \subfigure[]{\includegraphics[width=0.13\linewidth]{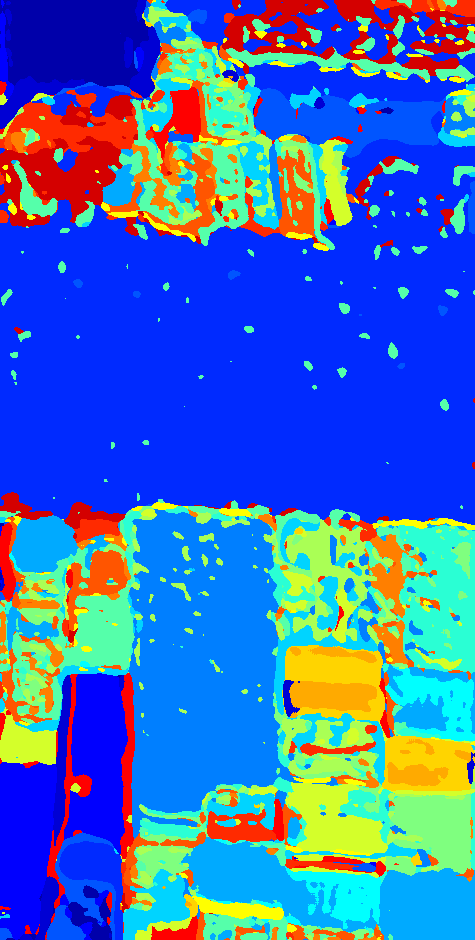}}
  \subfigure[]{\includegraphics[width=0.13\linewidth]{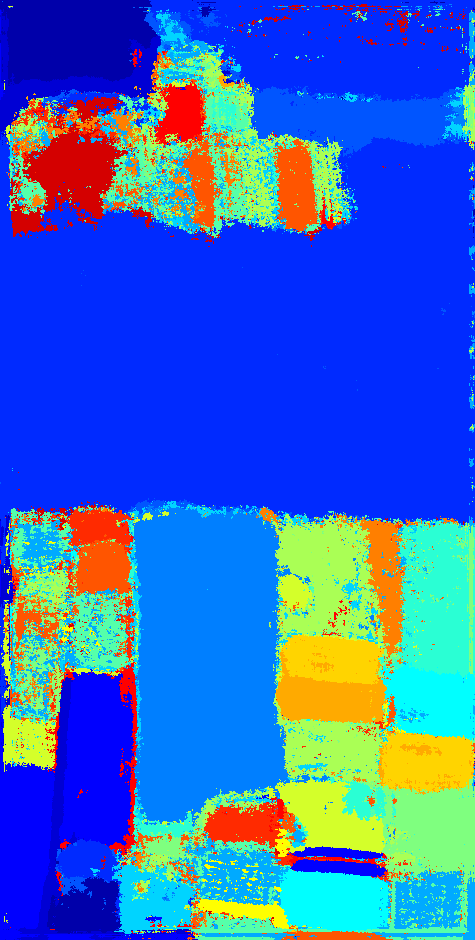}}
  \subfigure[]{\includegraphics[width=0.13\linewidth]{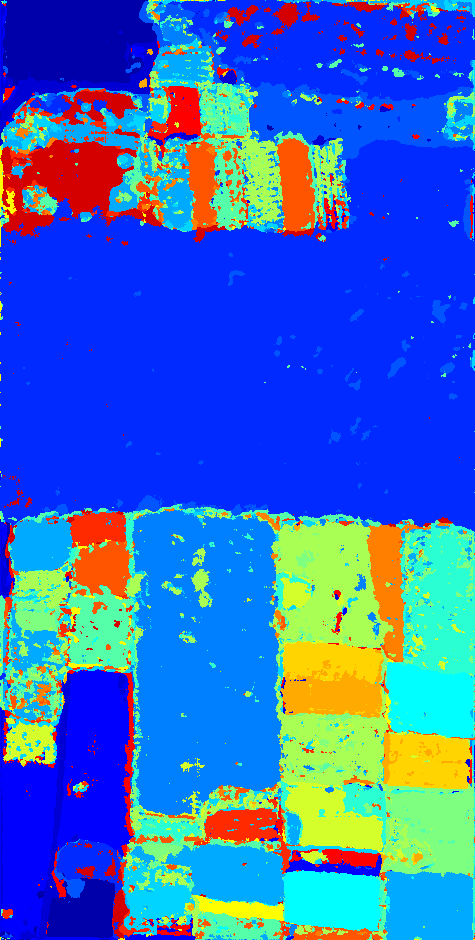}}
  \subfigure[]{\includegraphics[width=0.13\linewidth]{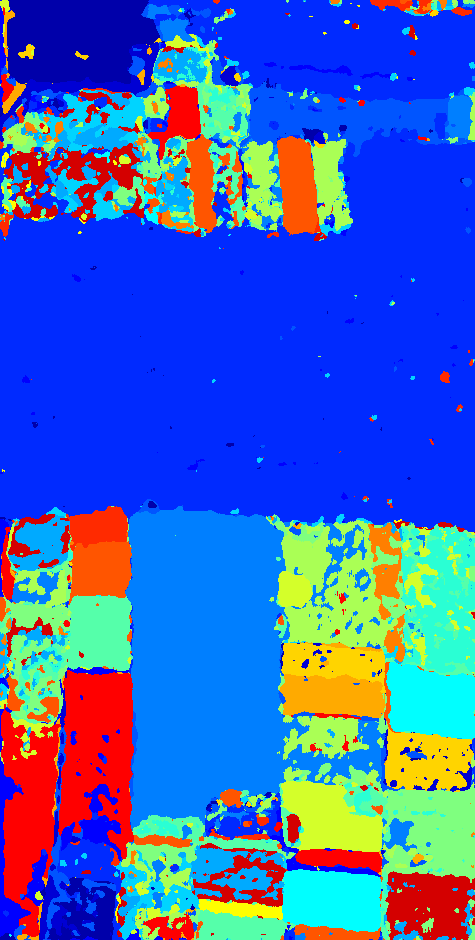}}
  \subfigure[]{\includegraphics[width=0.13\linewidth]{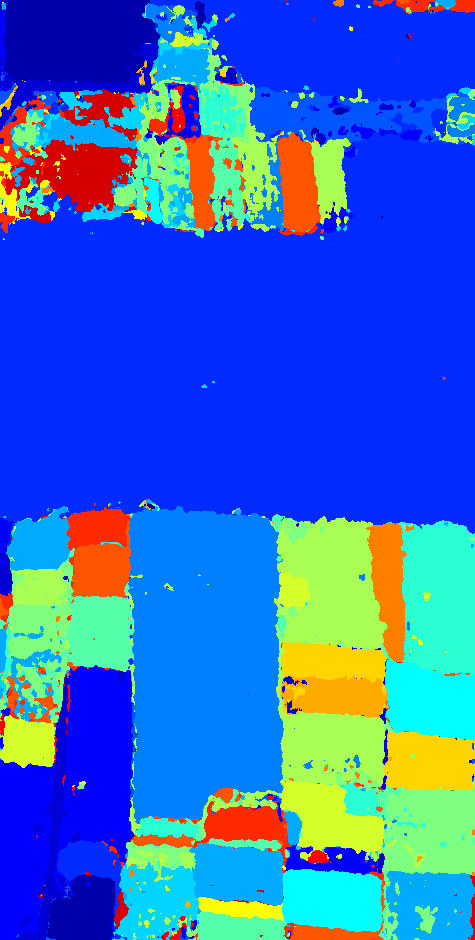}}
  \subfigure[]{\includegraphics[width=0.13\linewidth]{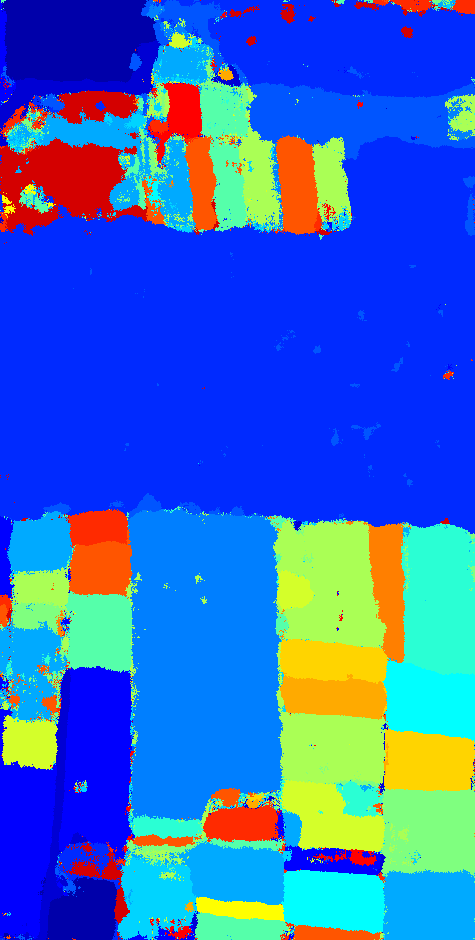}}\\
  \subfigure[]{\includegraphics[width=0.13\linewidth]{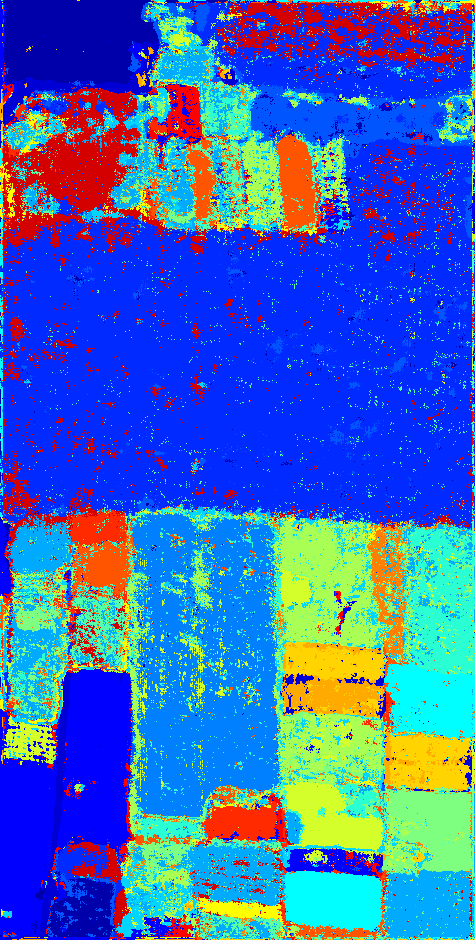}}
  \subfigure[]{\includegraphics[width=0.13\linewidth]{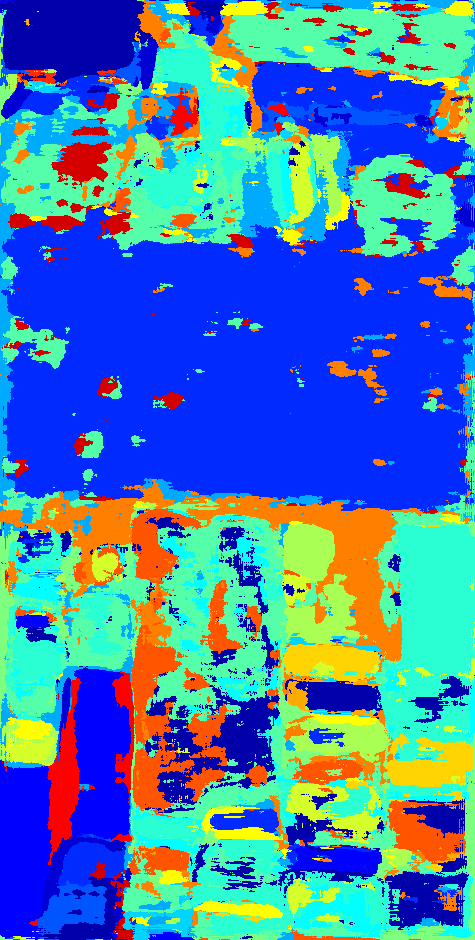}}
  \subfigure[]{\includegraphics[width=0.13\linewidth]{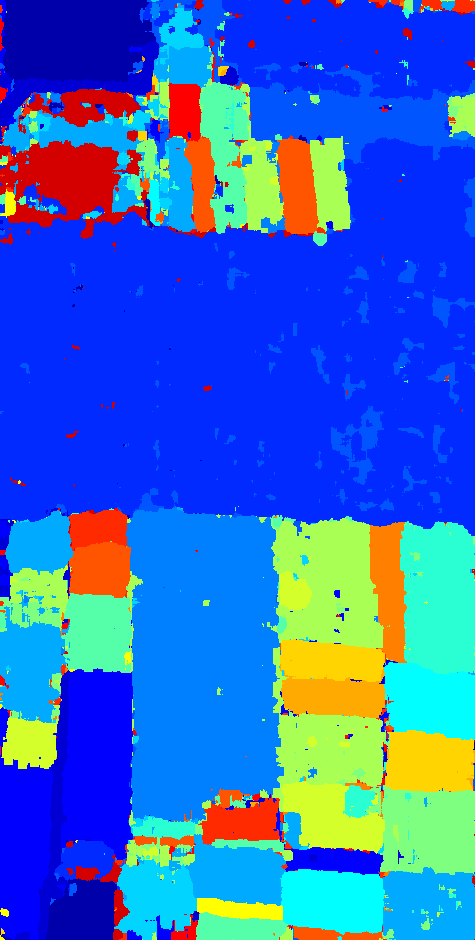}}
  \subfigure[]{\includegraphics[width=0.13\linewidth]{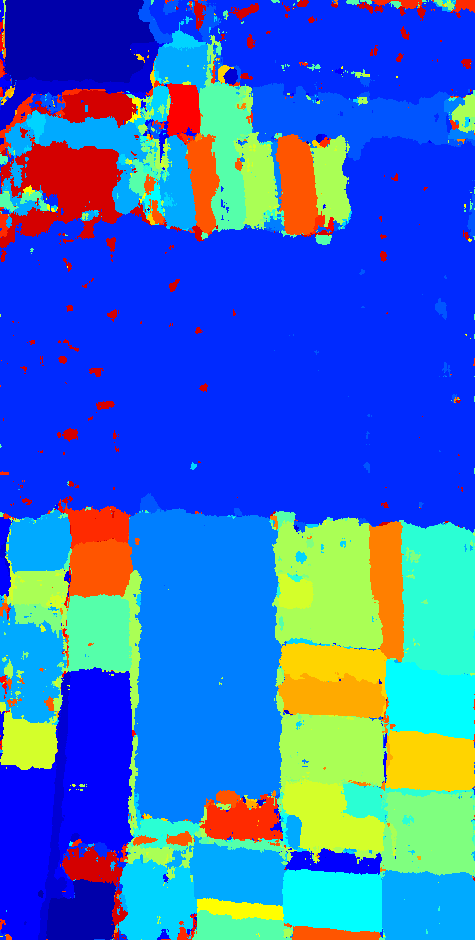}}
  \subfigure[]{\includegraphics[width=0.13\linewidth]{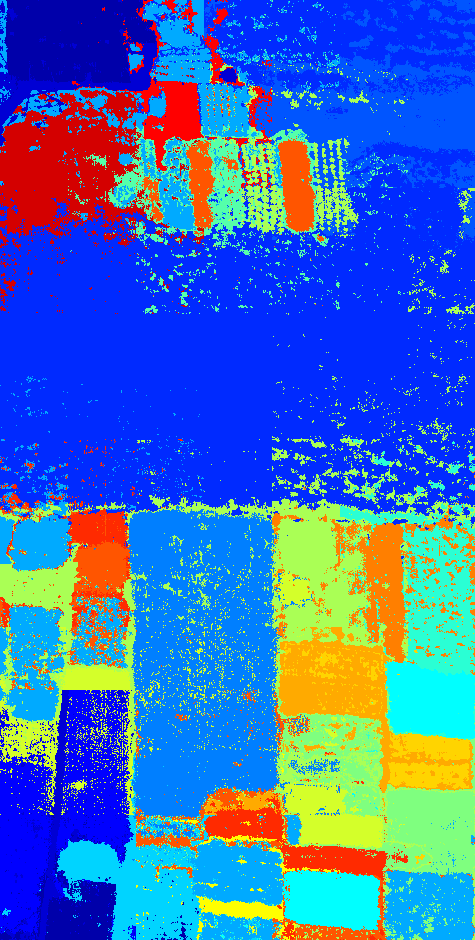}}
  \subfigure[]{\includegraphics[width=0.13\linewidth]{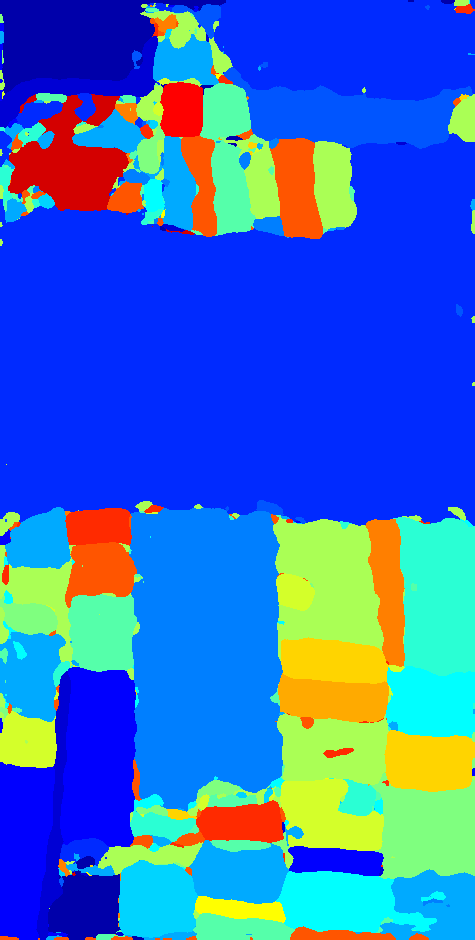}}
  \subfigure[]{\includegraphics[width=0.13\linewidth]{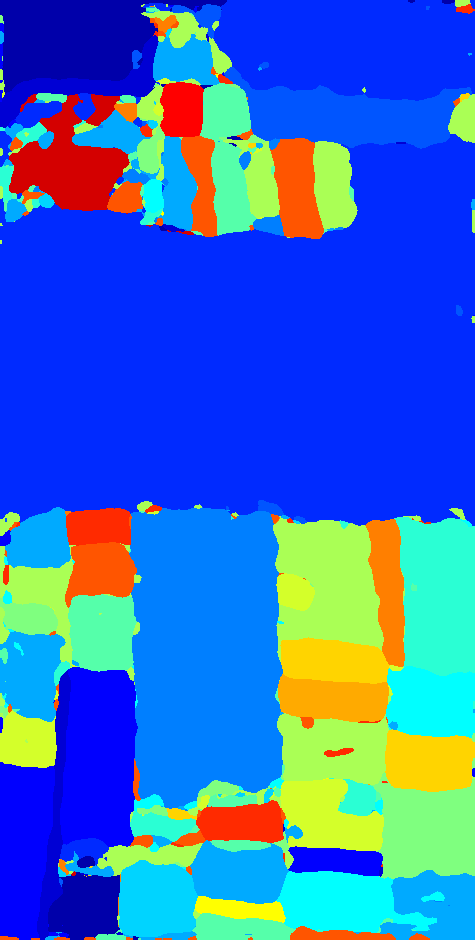}}
  \caption{Classification maps of different methods on the HongHu Scene in the WHU-Hi dataset (100 training samples per class). (a) Ground Truth. (b) 3DCNN. (c) SSAN. (d) SSFCN. (e) ENL-FCN. (f) WFCG. (g) HSI-BERT. (h) SpecFormer. (i) T-SST. (j) SSFTT. (k) LSFAT. (l) LESSFormer. (m) Ours (Hard Voting). (n) Ours (Soft Voting).}
  \label{honghu_cls}
\end{figure}

\begin{figure}[t]
  \centering
  \subfigure[]{\includegraphics[width=0.35\linewidth]{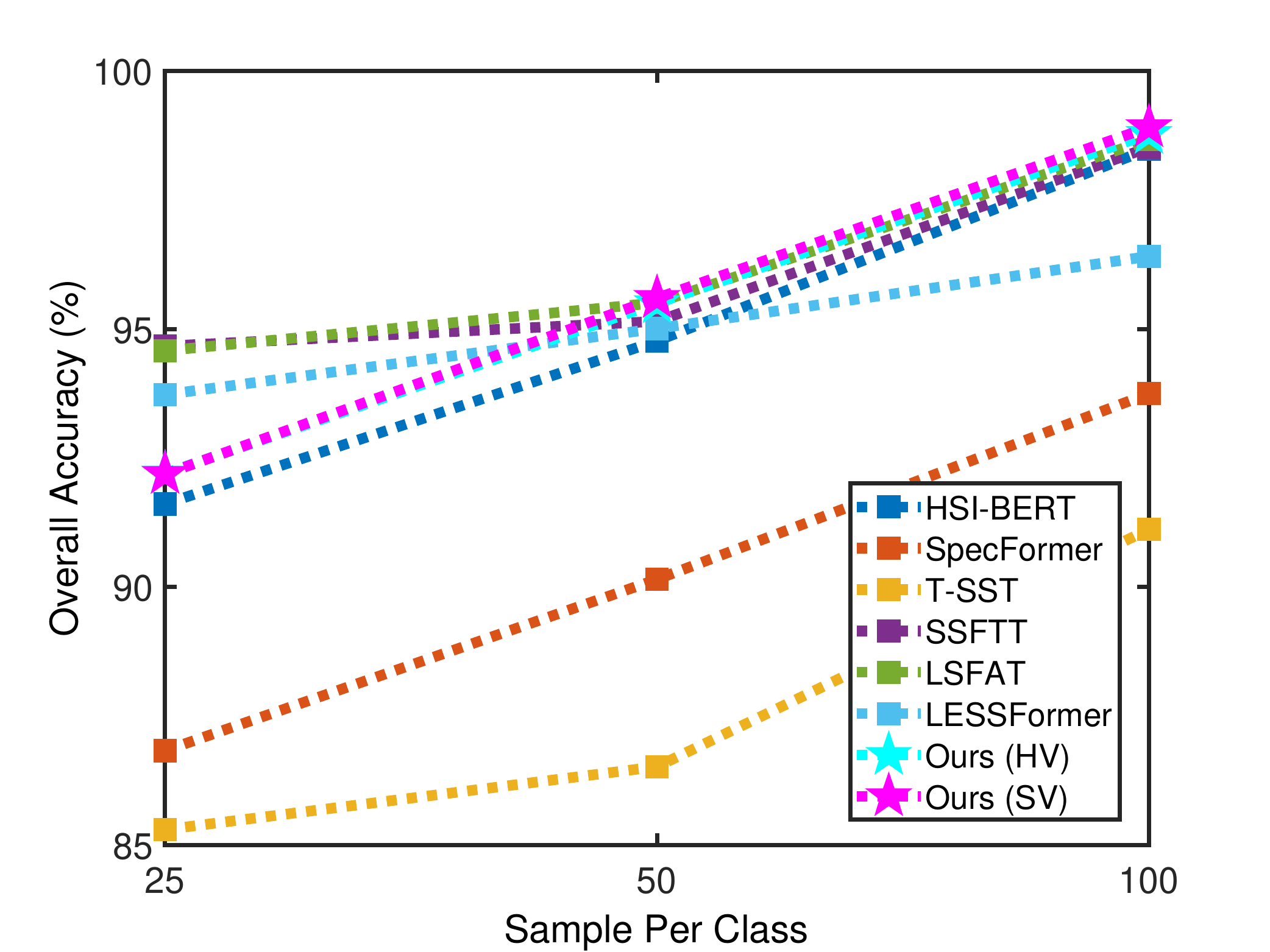}}
  \subfigure[]{\includegraphics[width=0.35\linewidth]{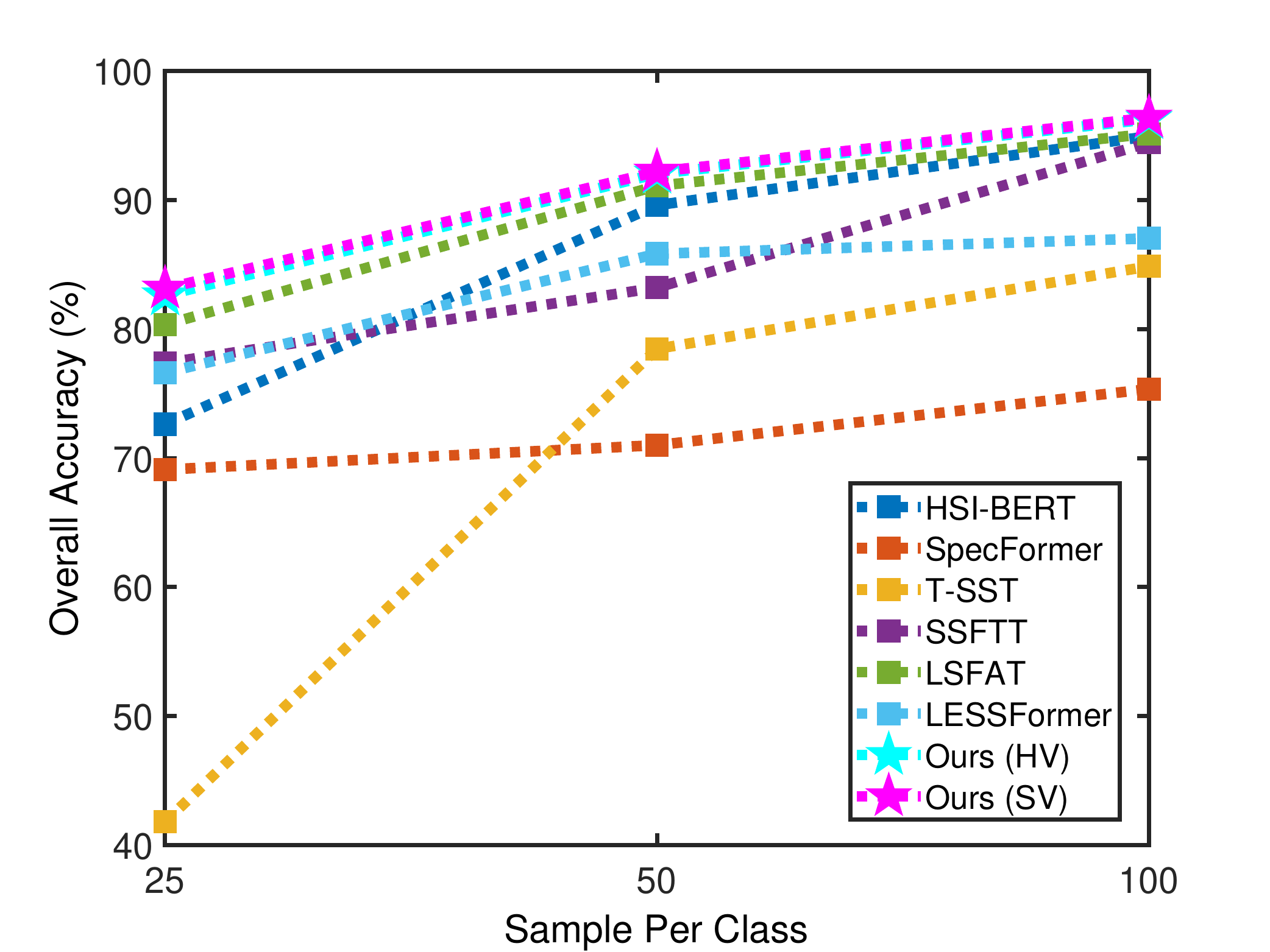}}\\
  \subfigure[]{\includegraphics[width=0.35\linewidth]{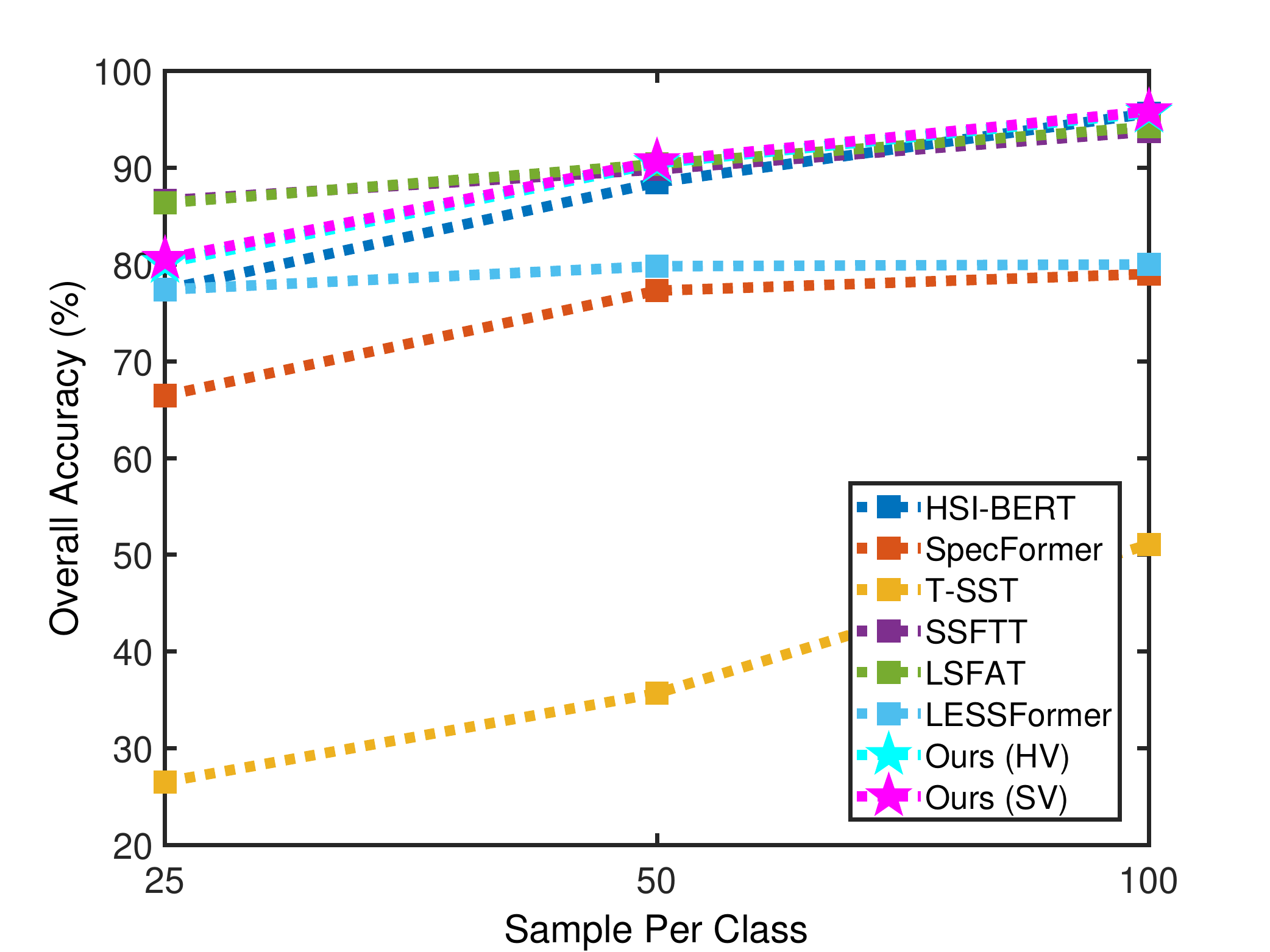}}
  \caption{The OA changes with different training samples for transformer-based methods on three scenes in the WHU-Hi dataset (Best viewed in color and zoom-in).}
  \label{less_sample}
\end{figure}

We compare our method with the existing state-of-the-art deep learning-based HSI classification approaches, including the pixel-level or patch-level methods such as 3-DCNN \cite{3dcnn} and SSAN \cite{ssan_rs}, image-level methods including SSFCN \cite{ssfcn}, ENL-FCN \cite{enl_fcn} and WFCG \cite{wfcg}, and recent transformer-based methods: HSI-BERT \cite{hsi_bert}, SpecFormer \cite{spectralformer}, T-SST \cite{tsst}, SSFTT \cite{ssftt}, LSFAT \cite{lsfat} and LESSFormer \cite{lessformer}. The implementation details of them are listed as follows:
\begin{itemize}
  \item [1)] In 3DCNN, 3-D convolutions are utilized to extract features from a 27 $\times$ 27 $\times$ $L$ cube ($L$ is the number of channels in HSI, which will be omitted if not specified) for classification. Following the default setting in the original paper, the network is trained by the SGD optimizer with a learning rate of 0.003.
  \item [2)] SSAN separately utilizes the input pixel vectors with $L$ channels and the 27 $\times$ 27 patches extracted from the first four principal components. The channel and position attentions are used to enhance them. We use the Adam optimizer with a learning rate of 0.005 to optimize the network.
  \item [3)] SSFCN is a two-branch FCN that processes a whole image. The final features are obtained by aggregating all deep and shallow features of these branches and used to predict a segmentation map. The network is trained by the SGD optimizer with a momentum of 0.9 and a learning rate of 0.001.
  \item [4)] ENL-FCN is a five-layer FCN and receives a whole image as the input. After the second block, two parallel non-local criss-cross SA modules are used to extract spatial contexts. We use the SGD optimizer with a learning rate of 0.001 to train it.
  \item [5)] WFCG also receives a whole image as the input. It uses two branches that separately contain full position SA and graph attention mechanism to extract different spatial features, which are then merged for classification. This network is trained by the Adam optimizer with a learning rate of 0.001.
  \item [6)] HSI-BERT uses the BERT \cite{bert} to extract bi-directional contexts from a flattened 11 $\times$ 11 patch. The network depth is 2. It uses 2 heads in the MHSA of the transformer encoder and the MLP ratio is 2. This network is trained by the AdamW optimizer with a learning rate of 0.0003.
  \item [7)] SpecFormer uses five cascade transformer encoders to capture spectral relationships in a 7 $\times$ 7 patch. The dimensions of embedding and the MLP hidden layers are 64 and 8. The number of SA heads is 4. This network is optimized by the Adam optimizer with a learning rate of 0.0005.
  \item [8)] T-SST processes 33 $\times$ 33 patches. Two transformer encoders are used to capture the relationships between the obtained CNN features. The number of SA heads is 2 and the MLP ratio is 4. We use the Adam optimizer with a learning rate of 0.0005 to train it.
  \item [9)] SSFTT extracts 13 $\times$ 13 patches from a PCA-compressed HSI that has 30 channels. A single-layer transformer encoder with 4 SA heads and an MLP ratio of 8 is used to process the flattened CNN feature maps. We use an Adam optimizer with a learning rate of 0.001 to train it.
  \item [10)] LSFAT processes a patch of size 15 $\times$ 15 $\times$ 30 from a PCA-compressed HSI with three transformer blocks. Each block has two transformer encoders with 4 MHSA heads and an MLP ratio of 2. This network is trained by the Adam optimizer with a learning rate of 0.001.
  \item [11)] LESSFormer processes a whole image by sequentially splitting the channels into four groups. It additionally introduces the division of numerous spatial regions. Then, four parallel transformer layers together with another transformer layer are separately used to extract features. Each layer has two transformer encoders. The head number is 4 and the MLP ratio is 2. We adopt the Adam optimizer with a learning rate of 0.0005 to train it.
\end{itemize}
Among the above methods, all of them are implemented by PyTorch except SSAN and HSI-BERT which are implemented by TensorFlow.

\noindent\textbf{Accuracy.} Table \ref{performance_compare} lists the comparison results, where the first and second places are marked by the bold and blue font, respectively. In pixel-level or patch-level CNN-based HSI classification methods such as 3DCNN and SSAN, the extracted feature is restricted by the input size, limiting the performance. Although SSFCN, ENL-FCN and WFCG leverage the whole image, they cannot effectively extract long-range context features due to the limited receptive field of the convolution. For the transformer-related methods, SpecFormer and T-SST attempt to use a transformer to directly model the relationships between adjacent bands. However, because of the spectral redundancies, it is difficult to capture the contexts between different channels, especially for the challenging WHU-Hi dataset. Although SSFTT and LSFAT consider extracting spatial features, their patch-based structure limits the range of context and affects the performance. We notice that the recently proposed LESSFormer simultaneously extracts image-level spatial and spectral features. Nevertheless, it only applies simple channel splitting and ignores the local context, and thus is unable to fully model both spatial and spectral relationships.

With the specially designed tri-spectral image generation pipeline, we address these issues and successfully encode the key spectral information into a series of tri-spectral images. The spectral information can be fully exploited through a voting scheme, where the ``soft'' scheme performs better since it uses the predicted probabilities of all classes. The obtained high-quality tri-spectral images make it possible to to adopt the ImageNet pretrained backbone networks to extract highly representative features, which are essential for the subsequent transformers to capture the two kinds of contexts effectively. Moreover, the design that our network processes the whole image also facilitates extracting effective contexts in the long range. Such a design is different from most existing transformer-based methods that are mainly implemented for pixel-level or patch-level classification. By sequentially extracting both intra- and inter-area spatial contexts, we obtain more discriminative spatial features. Thus, the proposed method performs the best on almost all metrics in these three scenes. Figure \ref{longkou_cls}-\ref{honghu_cls} presents the classification maps, where our method delivers fewer erroneous predictions and more clear region boundaries.

\begin{table}[t]
  \caption{The computational cost (s) of different transformer-based methods in the WHU-Hi dataset (100 training samples per class, counted on a single A100 GPU)}
  \newcommand{\tabincell}[2]{\begin{tabular}{@{}#1@{}}#2\end{tabular}}
  \centering
  \scriptsize
  \resizebox{\linewidth}{!}{
  \begin{tabular}{|l|cccccc|}
  \hline
  & \multicolumn{2}{c|}{LongKou} & \multicolumn{2}{c|}{HanChuan} & \multicolumn{2}{c|}{HongHu} \\ 
 \cline{2-7}
  \multirow{-2}{*}{Method} & $T_{trn}$ & $T_{tes}$ &  $T_{trn}$ & $T_{tes}$ &  $T_{trn}$ & $T_{tes}$ \\
  \hline
  HSI-BERT & 28.32 & 26.43 & 48.63 & 30.04 & 74.30 & 46.87 \\
  SpecFormer & 429.69 & 58.54 & 751.55 & 74.67 & 1033.87 & 111.29 \\
  T-SST & 87.72 & 78.78 & 135.27 & 95.86 & 188.87 & 155.87 \\
  SSFTT & 13.69 & 11.24 & 22.66 & 14.91 & 24.95 & 19.81 \\
  LSFAT & 29.98 & 22.43 & 43.87 & 25.06 & 58.11 & 39.12 \\
  LESSFormer & 15400.70 & 22.94 & 25044.29 & 36.67 & 31118.95 & 47.25 \\
  \hline
  Ours (HV) & 5248.48 & 82.92 & 5927.36 & 100.00 & 7339.78 & 119.20\\
  Ours (SV) & 5202.61 & 81.48 & 5935.76 & 94.38 & 7326.93 & 111.89 \\
  \hline
\end{tabular}
  }
  \label{compute_time}
\end{table}

\noindent\textbf{Fewer Samples.} We also compare our method with other transformer-based approaches by using different numbers of training samples, as shown in Figure \ref{less_sample}. It can be observed that patch-level methods e.g., SSFTT and LSFAT usually perform better when there are very few training samples. We suspect it is probably because the image-level methods are more prone to overfitting in this case due to the lack of sufficient mini-batches during training. While patch-level methods mitigate the issue by random sampling. Nevertheless, once the number of training samples increases, the potential of the image-level method can be fully exploited and the accuracies of the proposed method are rapidly improved. In addition, our method performs always the best on the challenging long strip scene HanChuan. It is probably because long-range context plays a more important role in this case. In other scenes, our method can deliver competitive performances when reducing training samples.

\noindent\textbf{Efficiency.} We compare the computational efficiency of different transformer-based methods, and the results are shown in Table \ref{compute_time}. It can be discovered that, compared with image-level methods, recent pixel-level or patch-level methods have less computational cost, since they usually utilize shallow layers, especially HSI-BERT, SSFTT, and LSFAT. However, their classification accuracies are also lower. Because of simultaneously adopting multiple transformer branches or blocks that contain transformer encoders to process the whole image, LESSFormer has the slowest running speed. Compared to the above models, our methods achieve a good trade-off between accuracy and efficiency, particularly in challenging scenes such as HanChuan and HongHu. In addition, we notice the running time of hard voting is slightly longer than soft voting. It is probably because the hard voting needs to generate prior classification maps $\widehat{Y}$ via $argmax$ in multiple times.

\begin{table}[t]
  \caption{The OAs (\%) with different backbones of the proposed method in the WHU-Hi dataset (100 training samples per class, HV: Hard Voting, SV: Soft Voting)}
  \newcommand{\tabincell}[2]{\begin{tabular}{@{}#1@{}}#2\end{tabular}}
  \centering
  \scriptsize
  \resizebox{\linewidth}{!}{
  \begin{tabular}{|l|c|cccccc|}
  \hline
  &  & \multicolumn{2}{c|}{LongKou} & \multicolumn{2}{c|}{HanChuan} & \multicolumn{2}{c|}{HongHu} \\ 
 \cline{3-8}
  \multirow{-2}{*}{Backbone} & \multirow{-2}{*}{Params. (M)} & HV & SV &  HV & SV &  HV & SV \\
  \hline
  VGG-16 & 138.4 & 98.78& 98.91 & 96.27 & 96.38 & 95.69 & 95.85 \\
  \hline
  ResNet-50 & 23.6 & 98.55 & 98.97 & 97.01 & 97.09 & 97.63 &  97.54 \\
  HRNet-W18-S-V1& 13.4 & 97.68 & 97.77 & 96.66& 96.78 & 96.85 & 96.82 \\
  MobileNet-V2-1.0 & 3.5 & 96.50 & 96.53 & 96.89 & 97.05 & 96.74 & 96.72 \\
  \hline
\end{tabular}
  }
  \label{different_backbone}
\end{table}

\noindent\textbf{Backbone.} Finally, we investigate the impact of the backbone network by evaluating the proposed method using a series of lightweight backbones, including ResNet-50 \cite{resnet}, HRNet-W18-S-V1 \cite{hrnet}, and MobileNet-V2-1.0 \cite{mobilenetv2}. The results presented in Table \ref{different_backbone} show that the advanced backbones proposed after VGG demonstrate better performance, particularly in challenging scenes such as HanChuan and HongHu, due to their modern designs such as the inverted residual block in MobileNet-V2, despite having fewer parameters. Overall, ResNet-50 outperforms the other backbones in all scenes owing to its strong representation capacity. These findings further support the effectiveness of DCN-T across different backbone networks.

\subsection{Visualization}

We visualize some intermediate results, including the generated partial tri-spectral images, internal features, as well as the training loss curves and test accuracy curves during network training to further demonstrate the effectiveness of our method.

\begin{figure}[t]
  \centering
  \includegraphics[width=0.9\linewidth]{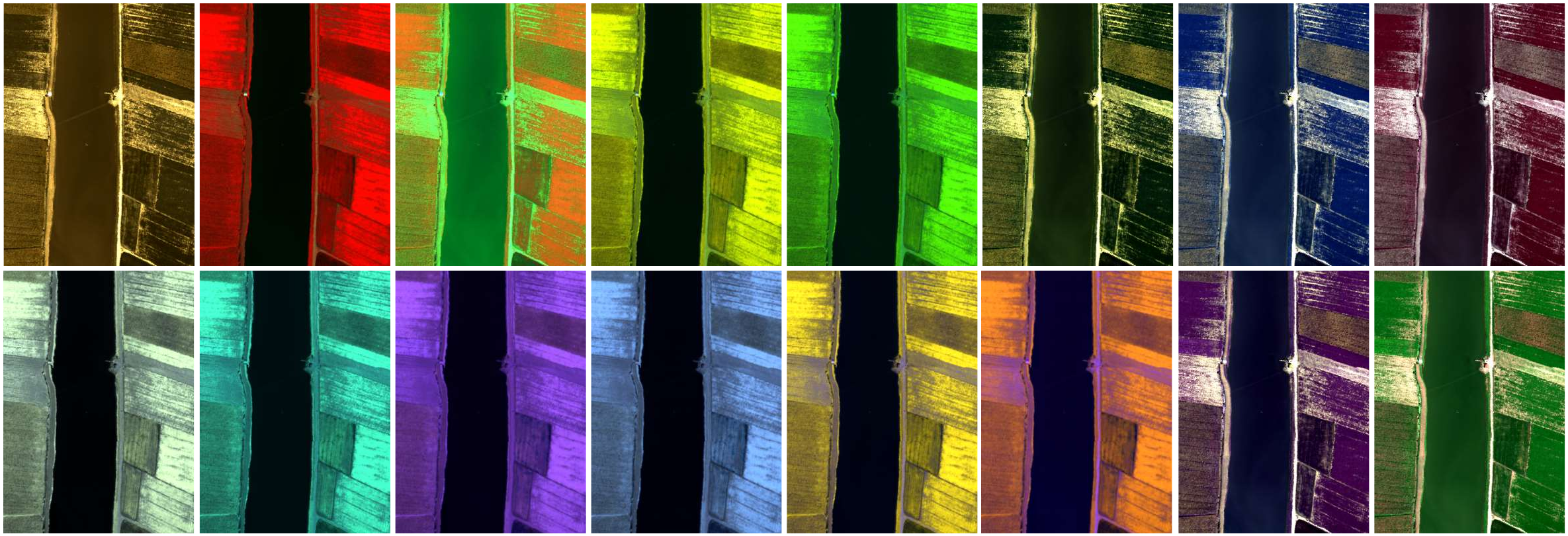}\\
  \caption{The obtained partial tri-spectral images on the LongKou Scene of the WHU-Hi dataset.}
  \label{rgb}
\end{figure}

\begin{figure}[t]
  \centering
  \includegraphics[width=0.7\linewidth]{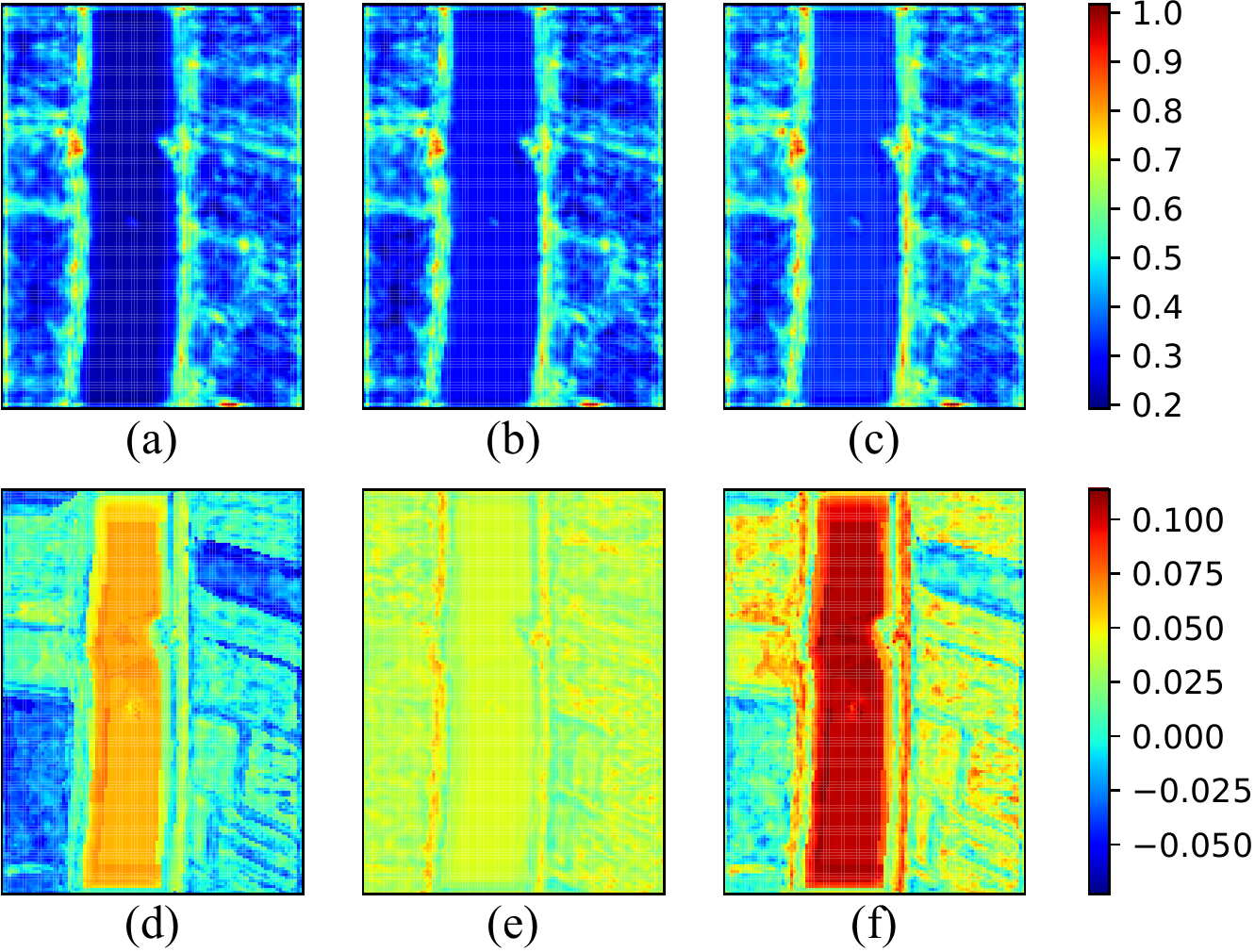}
  \caption{Visualization of internal features and their differences for the proposed network on the LongKou Scene in the WHU-Hi dataset, where each row shares the same color bar. (a) $\mathbf{F}$. (b) $\mathbf{F}^{RAC}$. (c) $\mathbf{F}^{GAC}$. (d) $\mathbf{F}^{RAC}-\mathbf{F}$. (e) $\mathbf{F}^{GAC}-\mathbf{F}^{RAC}$. (f) $\mathbf{F}^{GAC}-\mathbf{F}$}
  \label{features}
\end{figure}
\begin{figure}[t]
  \centering
  \subfigure[]{\includegraphics[width=0.48\linewidth]{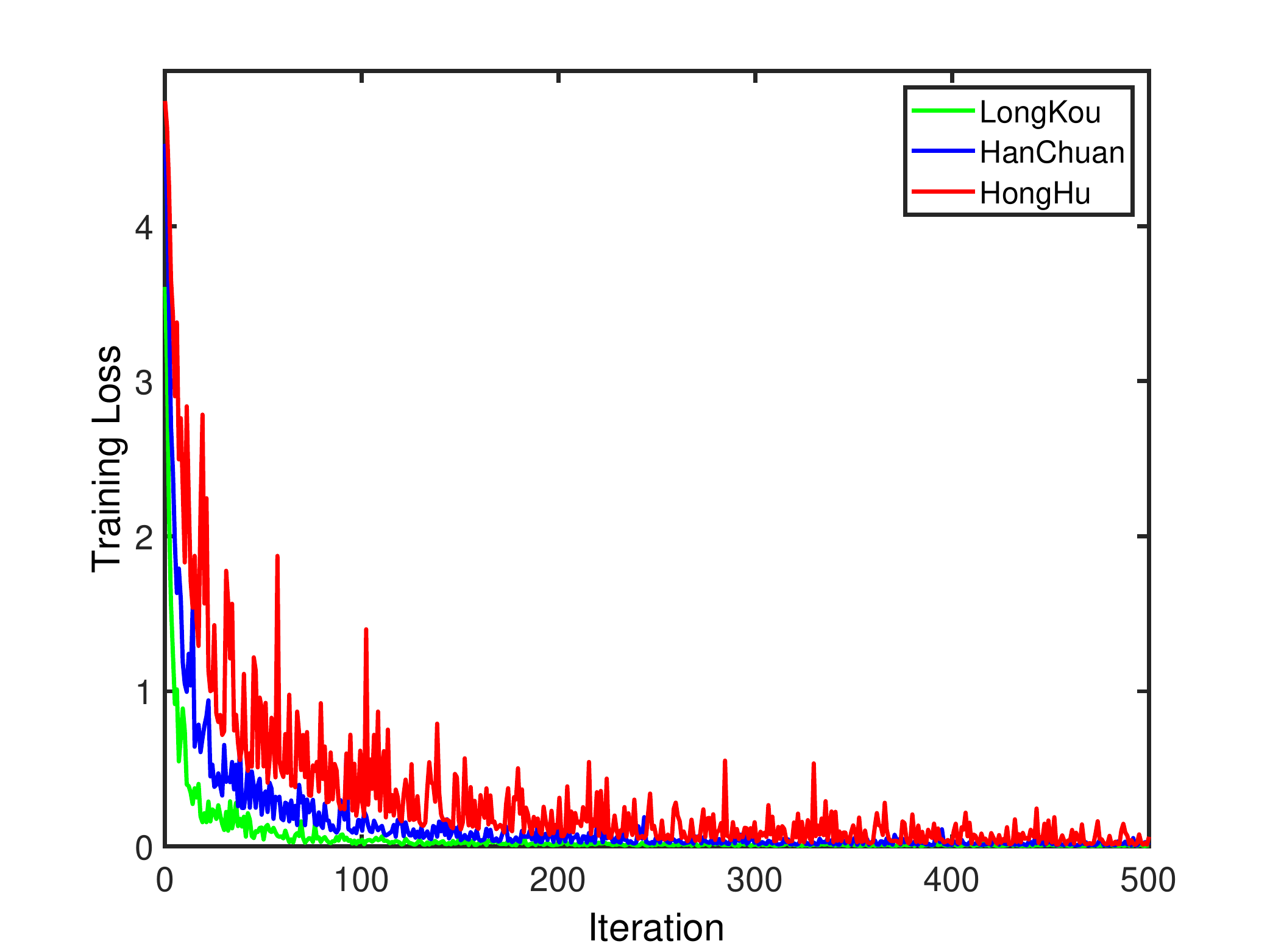}}
  \subfigure[]{\includegraphics[width=0.48\linewidth]{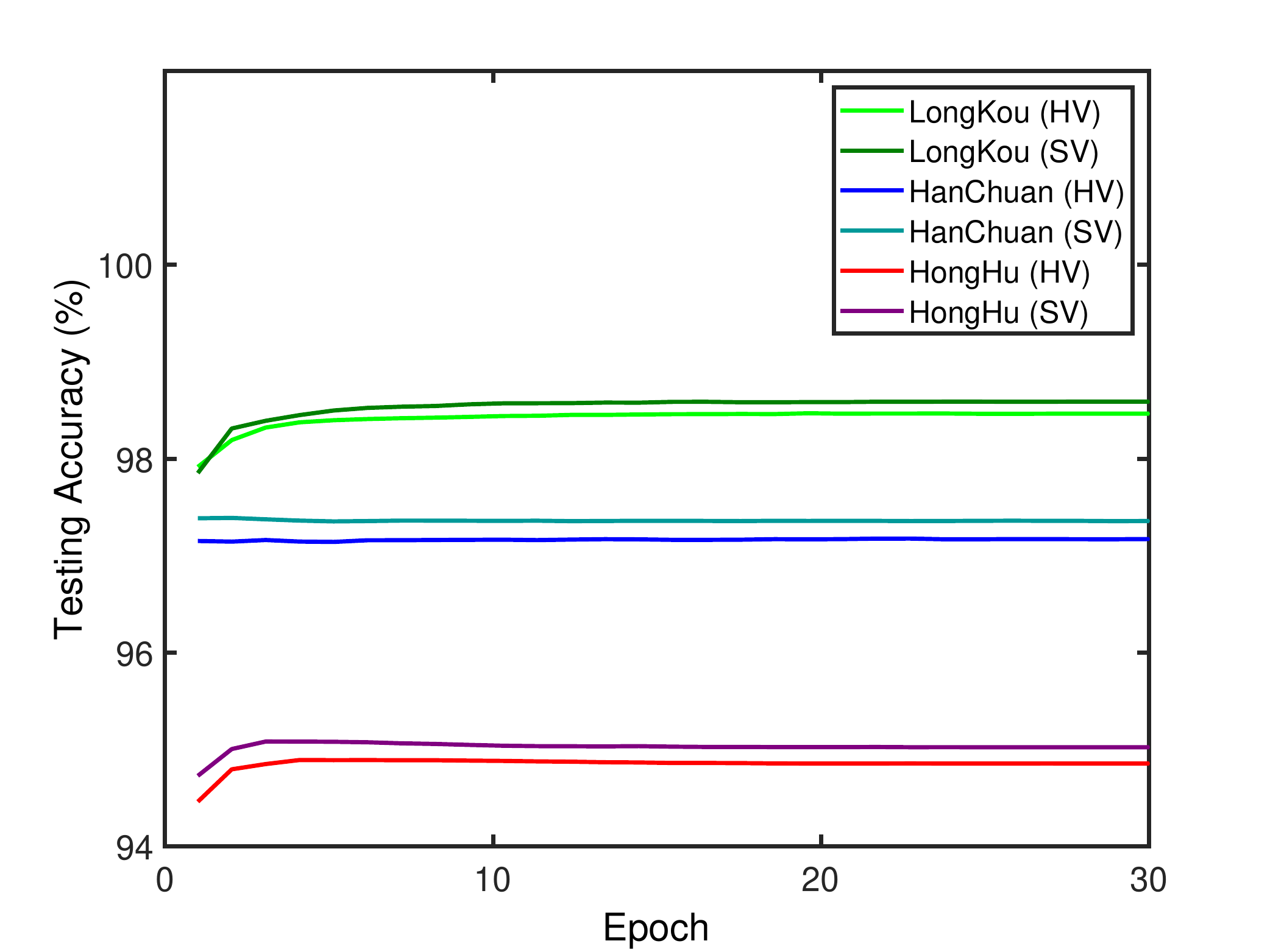}}
  \caption{Learning curves during network training on the WHU-Hi dataset (Best viewed in color and zoom-in). (a) Training loss. (b) Testing accuracy}
  \label{learn_curve}
\end{figure}

\noindent\textbf{Tri-spectral Dataset.} Figure \ref{rgb} depicts partial images of the generated tri-spectral dataset. It can be seen that the obtained tri-spectral images are colorful, where the current color reflects the corresponding spectral information that they carry. In fact, the aggregation operation removes spectral redundancy, while grouping and RBS diversify spectral information acquisition, to make full use of it. Thus, in our consideration, the proposed tri-spectral image generation pipeline optimizes the spectral information of the original HSI and transfers it to different tri-spectral images.

\noindent\textbf{Context Features.} We visualize the obtained $\mathbf{F}$, $\mathbf{F}^{RAC}$, $\mathbf{F}^{GAC}$ and their discrepancies in Figure \ref{features} to show the role of the DCM. Comparing (a), (b), and (c), it can be seen that the regional and global context modules improve features since the responsibility values are gradually increased. (d) depicts that the feature changes are presented by region, where most areas are boosted, especially for the river. (e) indicates that almost all locations in the scene are improved after global context capturing since each pixel representation has a global view. At last, (f) combines (d) and (e).

\noindent\textbf{Learning Curves.} Finally, we present the learning curves during network training in Figure \ref{learn_curve}. In the experiments, we observe that the network converges rapidly owing to ImageNet pretraining. Therefore, we only show the curves in the first 500 iterations for clarity. Moreover, we also plot the accuracy curves on the testing set evaluated after each epoch. It can be seen that the loss is decreased slower with more severe fluctuations in more challenging scenes, such as the HongHu. For the testing, it can be seen that the accuracies are rapidly saturated. We can also find that the performance of soft voting is slightly better than hard voting since the former leverages the predicted probabilities of all categories.

\section{Conclusion}

In this paper, we propose a novel approach named DCN-T for HSI classification. To deal with the high dimensionality issue caused by rich spectrums, we devise a tri-spectral image generation pipeline to transform the HSI into a series of high-quality tri-spectral images that possesses diverse refined spectral information. Then, the spatial variability of HSI is addressed by adopting an advanced ImageNet pretrained backbone network to extract highly representative features. Based on them, homogeneous areas that match object contours can be determined. Then, a novel dual context module based on the transformer is introduced to extract contexts within or between these areas, thereby improving feature representation and prediction accuracy. Finally, the prediction results of these tri-spectral images are combined using hard or soft voting schemes to further enhance performance. Experiments and ablation studies on three public benchmarks demonstrate the effectiveness of the proposed method and its superiority over state-of-the-art HSI classification methods.

\ifCLASSOPTIONcaptionsoff
  \newpage
\fi

\bibliographystyle{IEEEtran}
\bibliography{DCNT}

\end{document}